\documentclass[letterpaper]{article} 
\usepackage{aaai2026} 
\usepackage{times} 
\usepackage{helvet} 
\usepackage{courier} 
\usepackage[hyphens]{url} 
\usepackage{graphicx} 
\usepackage{natbib} 
\usepackage{caption} 
\usepackage[linesnumbered, ruled]{algorithm2e} 

\usepackage{newfloat}
\usepackage{listings}
\DeclareCaptionStyle{ruled}{labelfont=normalfont,labelsep=colon,strut=off} 
\usepackage{amsmath,amsfonts,bm}

\def\eqref#1{equation~\ref{#1}}

\def\1{\bm{1}}

\DeclareMathAlphabet{\mathsfit}{\encodingdefault}{\sfdefault}{m}{sl}
\SetMathAlphabet{\mathsfit}{bold}{\encodingdefault}{\sfdefault}{bx}{n}

\usepackage{amssymb}
\usepackage{amsmath}
\usepackage[capitalize]{cleveref}
\usepackage{threeparttable}
\usepackage{booktabs} 
\usepackage{multirow} 
\usepackage{makecell} 

\usepackage{colortbl} 
\usepackage{diagbox} 
\usepackage{siunitx}

\usepackage{enumitem} 
\usepackage{arydshln} 
\usepackage{tabularx}

\usepackage{xspace} 
\usepackage[subrefformat=parens, labelformat=parens]{
  subcaption
} 
\usepackage[Export]{adjustbox} 
\usepackage{pifont} 
\usepackage{xcolor}
\usepackage[dvipsnames]{xcolor}
\usepackage{placeins}

\crefname{table}{Tab.}{Tables.}
\crefname{figure}{Fig.}{Figures.}
\crefname{section}{Sec.}{Sections.}
\crefname{algorithm}{Algorithm }{Algorithms.}

\definecolor{mblue}{HTML}{367dbd}
\colorlet{colorFst}{mblue!45} 
\colorlet{colorSnd}{mblue!25} 
\colorlet{colorTrd}{yellow!30} 
\colorlet{colorLow}{darkgray!30} 
\definecolor{R1}{HTML}{E97451}
\definecolor{R2}{HTML}{008080}
\definecolor{R3}{HTML}{0047AB}
\colorlet{cmt}{darkgray!80} 
\colorlet{supp}{darkgray!50} 

\usepackage{environ}
\usepackage{xcolor}

\newcommand{\ours}{FreeGaussian\xspace}
\newcommand{\simdata}{OmniSim\xspace}
\newcommand{\realdata}{InterReal\xspace}
\newcounter{boldpara}
\newcommand{\boldparagraph}[1]{
\refstepcounter{boldpara}
\vspace{0.1em}
\noindent
{\bf #1} }

\newcommand{\rowhighlight}{\rowcolor{gray!30}}

\title{FreeGaussian: Annotation-free Control of Articulated Objects via \\
3D Gaussian Splats with Flow Derivatives}
\author{
  Qizhi Chen\textsuperscript{\rm 1, \rm 2}\equalcontrib,
  Delin Qu\textsuperscript{\rm 2, \rm 3}\equalcontrib,
  Junli Liu\textsuperscript{\rm 2, \rm 4},
  Yiwen Tang\textsuperscript{\rm 2},
  Haoming Song\textsuperscript{\rm 2}, \\
  Dong Wang\textsuperscript{\rm 2},
  Yuan Yuan\textsuperscript{\rm 4},
  Bin Zhao\textsuperscript{\rm 2, \rm 4}\thanks{Corresponding author.}
}
\affiliations{
  \textsuperscript{\rm 1}Zhejiang University
  \textsuperscript{\rm 2}Shanghai AI Laboratory
  \textsuperscript{\rm 3}Fudan University \\
  \textsuperscript{\rm 4}Northwestern Polytechnical University
}

\usepackage{bibentry}

\begin{document}
  \makeatletter \let\@oldmaketitle\@maketitle
  \renewcommand{\@maketitle}{\@oldmaketitle
  \begin{center}
    \captionsetup{type=figure}
    \setcounter{figure}{0}
    \includegraphics[trim=0ex 0 0 0, clip, width=0.95\textwidth]{
      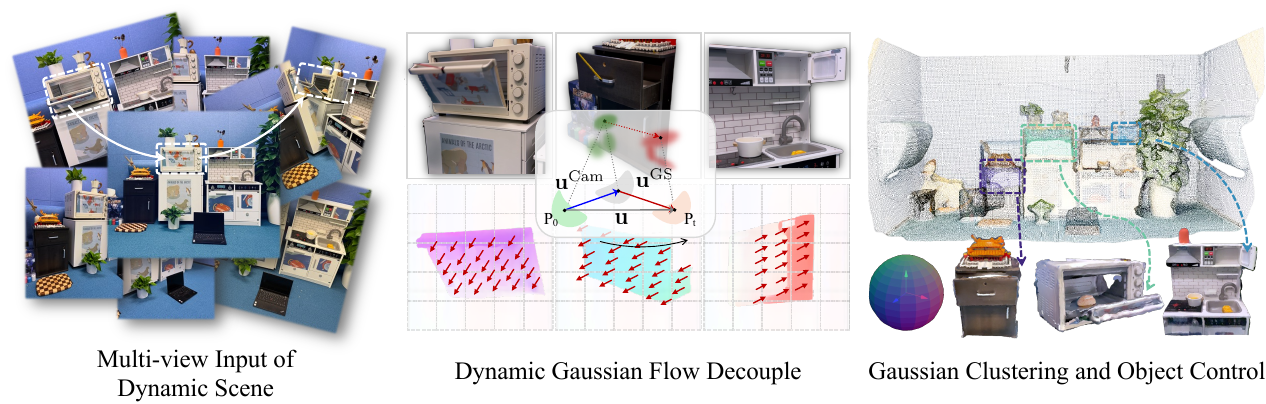
    }
    \caption{We present \ours, a novel annotation-free Gaussian Splatting method
    for controllable view synthesis, which connects optical flow, camera flow,
    and dynamic Gaussian flow through differential analysis. By refining Gaussian
    optimization with flow constraints, our approach improves motion smoothness,
    rendering quality, and eliminates manual annotations. Additionally, a 3D
    spherical vector control scheme simplifies interactive Gaussian modeling, demonstrating
    superior performance in view synthesis and individual object control. }
    \label{fig:teaser}
  \end{center}
  }
  \makeatother

  \maketitle

  \begin{abstract}
  Reconstructing controllable Gaussian splats for articulated objects from monocular video is especially challenging due to its inherently insufficient constraints. Existing methods address this by relying on dense masks and manually defined control signals, limiting their real-world applications. In this paper, we propose an annotation-free method, \textbf{\ours}, which mathematically disentangles camera egomotion and articulated movements via flow derivatives. By establishing a connection between 2D flows and 3D Gaussian dynamic flow, our method enables optimization and continuity of dynamic Gaussian motions from flow priors without any control signals. Furthermore, we introduce a 3D spherical vector controlling scheme, which represents the state as a 3D Gaussian trajectory, thereby eliminating the need for complex 1D control signal calculations and simplifying controllable Gaussian modeling. Extensive experiments on articulated objects demonstrate the state-of-the-art visual performance and precise, part-aware controllability of our method. Code is available at: 
  \textcolor{blue}{\url{https://github.com/Tavish9/freegaussian}}.
\end{abstract}

  \section{Introduction}
\label{sec:intro}
Controllable view synthesis (CVS) aims to recover scenes containing multiple articulated objects and interactable motions of each object given a set of views, it demands the recovered geometry, appearance, and motion faithfully respect the kinematic constraints of each articulated object while remaining photorealistic under novel viewpoints, which distinguishes it from conventional 4D reconstruction. Recently, CVS has attracted growing interest in content creation~\citep{liao2024advances,tang2023dreamgaussian,gao2024gaussianflow}, virtual reality~\citep{Steuer1992DefiningVR,kerbl20233d,Waisberg2023TheFO}, real-time reconstruction~\cite{qu2024implicit} and robotic manipulation~\citep{song2025hume,qu2025spatialvla,qu2025eo}. 

Recent advances leverage 3D Gaussian splatting~\citep{kerbl20233d} to achieve real-time, high-fidelity rendering of dynamic scenes~\citep{yu2023cogs,yang2023deformable3dgs} and have been scaled to scene-level datasets with dense annotations~\citep{Qu2024LiveSceneLE}. Yet, these methods remain fundamentally tied to manual supervision: they either require pixel-accurate part masks for each articulated link~\citep{yu2023cogs} or rely on pre-defined control signals in neural radiance fields~\citep{kania2022conerf, Qu2024LiveSceneLE}. Without mask or control signal supervision, the model collapses, failing to decode features to color and losing scene control capabilities. Thus, dense part masks and control signal annotations have become a prerequisite for current articulated-object CVS, severely limiting real-world deployment.

To address this challenge, we propose \textbf{\ours}, a annotation-free but effective Gaussian splatting method for controllable scene reconstruction, which automatically explores interactable structures and restores scenes from successive frames, without any manual annotations. Dynamic Gaussian flow under instantaneous motion can be analytically derived from optical flow and camera egomotion via differential analysis. It enables us to localize controllable structures without masks and estimates joint-angle trajectories without any control signals. These consistent constraints are folded into training, enabling high-fidelity rendering and fine-grained manipulation of articulated objects while eliminating the need for manual supervision and extending practical applicability to real-world scenes.

More specifically, in the training stage, \ours directly derive dynamic Gaussians flow from optical flow and camera-induced camera flow, accumulated with Gaussian projection displacements. By tracking the dynamic Gaussian flow, we highlight interactive dynamic Gaussians and obtain their trajectories via HDBSCAN clustering, eliminating the dependence on manual mask annotations. To overcome the reliance on 1D control signal inputs, we introduce a 3D spherical vector controlling scheme that exploits 3D Gaussian scene representations bypassing dynamic Gaussian trajectories as state representations, aligning with the splatting rasterization pipeline and greatly simplifying the control process. During the control stage, the Gaussian dynamics are retrieved from the network, given the 3D control vector as input. Beyond localizing interactive Gaussians, the dynamic Gaussian flow constraints 3DGS motion between frames, guaranteeing smooth motion and eliminating ghosting artifacts to improve rendering quality.

Extensive evaluations show that our method outperforms existing methods significantly in both novel view synthesis and articulated object controlling, enabling more accurate and efficient modeling of interactable content with no annotations. Contributions can be summarized as follows:

\begin{itemize}[noitemsep, topsep=10pt, leftmargin=10pt]
  \item We propose \textbf{\ours}, a novel annotation-free Gaussian Splatting method for controllable scene reconstruction, which automatically explores interactable scene objects with flow priors, and restores scene interactivity without any manual annotations.

  \item \ours analytically derive the \textbf{dynamic Gaussian flow constraints} via differential analysis with alpha composition, which draws the mathematical link among optical flow, camera motion, and dynamic Gaussian flow. The flow constraints refine Gaussian optimization enabling unsupervised interactive structure localization and the training of continuous Gaussian motion variations.

  \item Exploiting 3D Gaussian explicitness, we introduce a \textbf{3D spherical vector controlling scheme}, avoiding traditional complex 1D control variable calculations bypassing 3DGS trajectory as state representation, further simplifying and accelerating interactive Gaussian modeling.

\end{itemize}
  \section{Related Work}

\label{sec:relate}
\subsubsection{4D Novel View Synthesis.} 
Neural Radiance Fields (NeRF)~\citep{mildenhall2020nerf} has innovated great progress in dynamic scene reconstruction. The existing methods can be categorized into three primary categories: time-varying methods~\citep{du2021neural,fang2022fast,li2021neural,park2021nerfies,pumarola2021d,tretschk2021non,yuan2021star} that append temporal embeddings and scene-flow to the radiance MLP; deformable-canonical approaches~\citep{gao2021dynamic,li2022neural,park2021hypernerf,xian2021space,wang2025monofusion} warp query points from a dynamic space to a static canonical volume; and hybrid representations~\citep{shao2023tensor4d,kplanes_2023,Cao2023HexPlane,song2023nerfplayer} have accelerated training and rendering via time-space feature planes, dynamic voxels, or 4D hash encodings. More recently, 3D Gaussian Splatting (3DGS)~\citep{kerbl20233d} has gained prominence due to its superior training efficiency and real-time rendering. Subsequent 3DGS extensions for dynamic scenes learn dense Gaussian trajectories directly~\citep{yang2023deformable3dgs,luiten2023dynamic}, augmenting 3DGS with 4D feature planes~\citep{wu20234dgaussians} or learnable motion bases~\citep{kratimenos2024dynmf}, and incorporating flow-based regularisation losses to enforce temporal consistency. 

\subsubsection{Controllable Scene Representation.}Decoupling appearance, geometry, and time has unlocked controllable avatars~\cite{Rivero2024Rig3DGSCC,liu2023humangaussian} and interactive simulators~\cite{Qu2024LiveSceneLE,Wang2024NeRFIR}. CoNeRF~\citep{kania2022conerf} pioneered this effort by extending HyperNeRF~\citep{park2021hypernerf} and regressing the attribute and the mask to enable few-shot attribute control. CoGS~\citep{yu2023cogs} leveraged 3D Gaussians to achieve real-time control of dynamic scenes without requiring explicit control signals. LiveScene~\citep{Qu2024LiveSceneLE} scales to scene level via factorized interactive space. But all these methods remain limited by dense manual annotations. More recently, MotionGS~\citep{zhu2025motiongs} explores explicit motion priors to guide the deformation of 3D Gaussian.

\begin{figure*}[t]
    \begin{center}
        \includegraphics[width=1.0\linewidth]{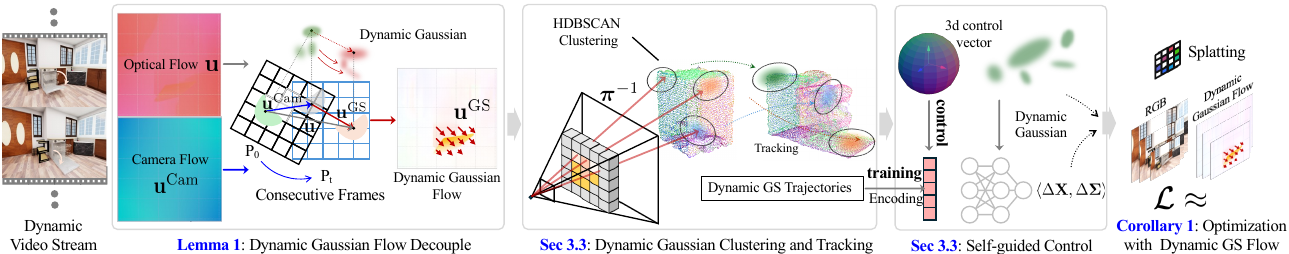}
    \end{center}
    \caption{
        The overview of \ours. Given a set of video stream $\{\mathbf{P}(t), \mathbf{I}(t)\}$, our method recovers controllable 3D Gaussians $\mathbf{G}^{\ast}$ with two stages. First, we pre-train a deformable 3DGS and calculate dynamic Gaussian flow $\mathbf{u}^\text{GS}$ via~\cref{eq:gaussian_flow_analysis}. Then, we reproject dynamic Gaussian flow maps and cluster the active Gaussians with HDBSCAN algorithm, followed by trajectory calculation. In the controllable training stage, we optimize Gaussians $\mathbf{G}$ and network $\mathbf{\Theta}$ under the rasterisation loss in~\cref{eq:loss}, which jointly aligns rendered images with input views and enforces consistency in the predicted dynamic flows.
    }
    \label{fig:pipeline}
\end{figure*}

  \section{Methodology}
\label{sec:method}

As depicted in \cref{fig:pipeline}, our approach exploits the underlying
connections among dynamic Gaussian flow, optical flow, and camera motion to
achieve annotation-free interactive scene reconstruction. The dynamic Gaussian flow
autonomously segments interactable objects, forming the basis for downstream
articulated object control. This enables trajectory-guided clustering and integrates
with a 3D spherical vector control framework, resulting in a streamlined and
scalable Gaussian modeling pipeline for dynamic scenes.

We first review 3DGS basics in \cref{sec:preliminary}, then formulate the connection
between optical flow, camera motion, and dynamic Gaussian flow in \cref{sec:differential_analysis}.
Based on this, we introduce a 3D spherical vector control scheme in \cref{sec:self_guided_control},
which discovers and clusters dynamic Gaussians via trajectory analysis. The full
pipeline is optimized with joint loss functions detailed in \cref{sec:loss_functions}.

\subsection{Preliminary of 3DGS Rasterization}
\label{sec:preliminary} 3D Gaussian Splatting~\citep{kerbl20233d} explicitly
represents scenes with millions of Gaussians and emerges ultra high-quality
rendering performance recently. Given a set of images capture with corresponding
camera poses, 3DGS models scenes by learning a set of 3D Gaussians
$\mathbf{G}= \{G_{i}: (\mathbf{X}_{i}, \mathbf{\Sigma}_{i}, \mathbf{o}_{i}, \mathbf{H}
_{i}) | i=1,...,N \}$, where $\mathbf{X}_{i}\in \mathbb{R}^{3}$,
$\mathbf{\Sigma}_{i}\in \mathbb{R}^{3 \times 3}$, $\mathbf{o}_{i}\in \mathbb{R}$,
and $\mathbf{H}_{i}\in \mathbb{R}^{48}$ are the center position, 3D covariance,
opacity, and spherical harmonics of the $i$-th Gaussian, respectively. With the rasterization
pipeline, 3DGS projects $\mathbf{G}$ to image planes as 2D Gaussians
$\mathbf{g}= \{g_{i}:(\boldsymbol{\mu}_{i}, \mathbf{\Sigma}_{i}^{\prime}, \mathbf{o}
_{i}, \mathbf{c}_{i}) | i=1,...,N\}$
and blender pixel colors $\hat{\mathbf{C}}$ via alpha composition:
\begin{equation}
	\hat{\mathbf{C}}= \sum_{i=1}^{N}\mathbf{c}_{i}\alpha_{i}T_{i},\quad T_{i}= \prod
	_{j=1}^{i-1}(1-\alpha_{j}), \label{eq:render}
	\vspace{-1ex}
\end{equation}
where $\boldsymbol{\mu}_{i}\in \mathbb{R}^{2}$ , $\mathbf{\Sigma}_{i}^{\prime}\in
\mathbb{R}^{2 \times 2}$, $\mathbf{c}_{i}\in \mathbb{R}^{3}$, $\alpha_{i}\in [0,1
]$ and $T_{i}\in [0,1]$ are the 2d center, 2d covariance, color, alpha value and
transmittance of 2D Gaussian $g_{i}$. The alpha value $\alpha_{i}$ at pixel coordinate
$\mathbf{m}$ can be obtained by:
\begin{equation}
	\alpha_{i}= \mathbf{o}_{i}\exp(-\frac{1}{2}(\mathbf{m}-\boldsymbol{\mu}_{i})^{T}
	\mathbf{\Sigma}_{i}^{\prime-1}(\mathbf{m}-\boldsymbol{\mu}_{i})). \label{eq:alpha}
\end{equation}
With the supervision of observations, 3DGS optimizes parameters to minimize the photometric
loss between rendered and ground-truth images.

\begin{figure}[t]
	\begin{center}
		\includegraphics[width=0.95\linewidth]{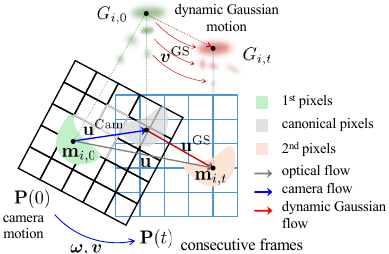}
	\end{center}
	\caption{Dynamic Gaussian flow illustration. In interactive scenes,
	consider an instantaneous motion model, where the camera and 3D Gaussian hold
	separate velocities in consecutive frames. The projected optical flow
	$\mathbf{u}$ can be decomposed into camera flow $\mathbf{u}^{\text{Cam}}$
	and dynamic Gaussian flow $\mathbf{u}^{\text{GS}}$, as described in~\cref{eq:gaussian_flow_analysis,eq:dynamic_gs_flow}.}
	\label{fig:dynamic_gaussian}
\end{figure}

\begin{figure*}[t]
	\begin{center}
		\includegraphics[width=1.0\linewidth]{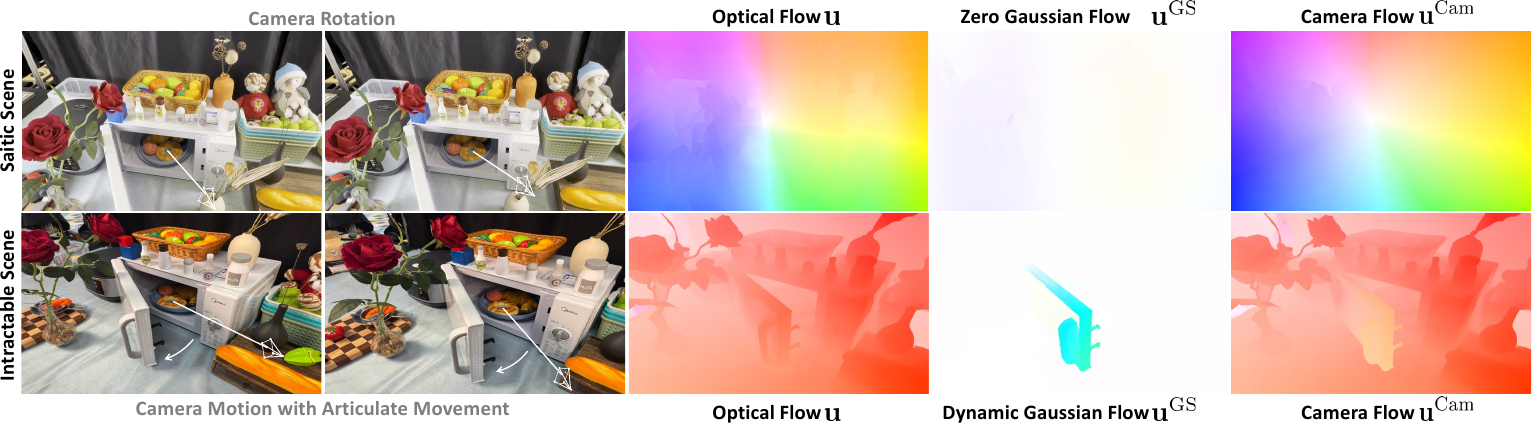}
	\end{center}
	\caption{ Illustration of dynamic Gaussian flow map under static and dynamic
	scenes. a) In static scenes with solely camera motion, ~\cref{eq:dynamic_gs_flow}
	degenerate to pure camera flow, yielding zero dynamic Gaussian flow. b) In
	contract, when articulated object moves, the dynamic Gaussian flow map will
	highlight interactive 3D Gaussians. }
	\label{fig:flowmap}
\end{figure*}

\subsection{Dynamic Gaussian Flow Analysis}
Our insight is that the dynamic Gaussian flow under instantaneous motion can be analytically
decoupled from optical flow and camera motion via differential analysis with alpha
composition. Considering a dynamic scene with interactive objects as shown in~\cref{fig:dynamic_gaussian},
the camera and 3D Gaussians hold separate velocities in consecutive frames $0$ and
$t$. Assuming a dynamic 3D Gaussian $G_{i}$ with velocity
$\boldsymbol{v}^{\text{GS}}$, it is projected as image measurement $g_{i}$ under
the constant camera instantaneous motion by translation velocity $\boldsymbol{v}$
and rotational velocity $\boldsymbol{\omega}$. The optical flow $\mathbf{u}$
induced by $(\boldsymbol{v}, \boldsymbol{\omega})$ of a pixel
$\mathbf{m}= (x,y)^{\top}$ can be obtained by \emph{Lemma 1}:

\label{sec:differential_analysis}

\subsubsection{Lemma 1:}
\textit{Dynamic Gaussian flow $\mathbf{u}^{\text{GS}}$ under instantaneous
motion can be derived from optical flow $\mathbf{u}$ and camera flow
$\mathbf{u}^{\text{Cam}}$ with the following transform~\cref{eq:gaussian_flow_analysis}.}
\begin{equation}
	\small
	\begin{aligned}
		\label{eq:gaussian_flow_analysis} & \mathbf{u}= \mathbf{u}^{\text{Cam}}+ \mathbf{u}^{\text{GS}}+ \mathbf{\Delta}, \quad \mathbf{u}^{\text{Cam}}= \frac{\mathbf{A}\boldsymbol{v}}{Z}+ \mathbf{B}\boldsymbol{\omega},                                               \\
		\\
		                                  & \mathbf{u}^{\text{GS}}= \mathbf{A}\sum_{i=1}^{M}T_{i}\alpha_{i}\frac{\boldsymbol{v}^\text{GS}}{Z_i}, \mathbf{\Delta}= \mathbf{A}\sum_{i=1}^{M}T_{i}\alpha_{i}\boldsymbol{v}(\frac{1}{Z_i}- \frac{1}{Z}),                      \\
		                                  & \mathbf{A}= \begin{bmatrix}-f_{x}&0&x - c_{x}\\ 0&-f_{y}&y - c_{y}\end{bmatrix},                                                                                                                                              \\
		                                  & \mathbf{B}= \begin{bmatrix}\frac{(x - c_x)(y - c_y)}{f_y}&- f_{x}- \frac{(x - c_x)^2}{f_x}&\frac{(y - c_y) f_x}{f_y}\\ f_{y}+ \frac{(y - c_y)^2}{f_y}&-\frac{(x - c_x)(y - c_y)}{f_x}&-\frac{(x - c_x)f_y}{f_x}\end{bmatrix}.
	\end{aligned}
\end{equation}
where $f_{x}, f_{y}, c_{x}, c_{y}$ are camera intrinsics, $M$ denotes the number
of Gaussian projections sorted with Gaussian depth $Z_{i}$ intersecting the
pixel $\mathbf{m}$. Flow residual term $\mathbf{\Delta}$ are preserved to guarantee
accuracy, even when it approaches zero after refined optimization. The proof involves
analyzing camera motion and dynamic GS motion under instantaneous motions, which
are detailed in \textbf{supplementary~\cref{sec:supp_differential_analysis}}.


The expression~\cref{eq:gaussian_flow_analysis} elucidates the triadic
relationship, yet Gaussian flow is not amenable to joint 3DGS training. For
flexibility, we consider a pixel $\mathbf{m}_{i,t}$ following 2D Gaussian distribution
$g_{i}$ at time $t$, and obtain
$\mathbf{m}_{i,t}\sim \mathcal{N}(\boldsymbol{\mu}_{i,t}, \mathbf{\Sigma}^{\prime}
_{i,t})$, with 2D mean $\boldsymbol{\mu}_{i,t}$ and covariance $\mathbf{\Sigma}_{i,t}
^{\prime}= \mathbf{B}_{i,t}\mathbf{B}_{i,t}^{\top}$. The following \textit{Corollary}
describes the dynamic Gaussian flow with 2D Gaussian means.

\vspace{1ex}
\noindent
\textbf{Corollary 1:}
\textit{The dynamic Gaussian flow $\mathbf{\tilde{u}}^{\text{GS}}$ on image
plane can be accumulated with 2D Gaussian means displacement
$\boldsymbol{\mu}_{i,t}- \boldsymbol{\mu}_{i,0}$.}
\begin{equation}
	\begin{aligned}
		\label{eq:dynamic_gs_flow}\mathbf{u}= \mathbf{u}^{\text{Cam}}+ \tilde{\mathbf{u}}^{\text{GS}}+ \mathbf{\Delta}, \\
		\tilde{\mathbf{u}}^{\text{GS}}= \sum_{i=1}^{M}T_{i}\alpha_{i}(\boldsymbol{\mu}_{i,t}- \boldsymbol{\mu}_{i,0}).
	\end{aligned}
\end{equation}
Detailed proof can be found in \textbf{supplementary~\cref{sec:supp_differential_analysis}}.

\subsubsection{Discussion.}
The expression in~\cref{eq:gaussian_flow_analysis,eq:dynamic_gs_flow} reveals dynamic
gaussian flow can be directly derived from 2D image flow $\mathbf{u}$ and camera-induced
camera flow $\mathbf{u}^{\text{Cam}}$, accumulated with 2DGS projection
displacement $\boldsymbol{\mu}_{i,t}- \boldsymbol{\mu}_{i,0}$. This naturally aligns
with the 3D Gaussian rasterization pipeline, providing continuous motion
constraints for dynamic Gaussian optimization. Besides, in static Gaussian
scenes, the equation degenerates to camera flow with $\mathbf{u}= \mathbf{u}^{\text{Cam}}$.
Hence, the resulting dynamic Gaussian flow map will highlight interactive 3D Gaussians,
as illustrated in~\cref{fig:flowmap}.

\noindent
Compared with GaussianFlow~\citep{gao2024gaussianflow}, which lacks explicit
camera motion modeling, and MotionGS~\citep{zhu2025motiongs}, which relies on back-projection
from known camera poses, our method is more general and flexible, benefiting from
a principled formulation under instantaneous motion.

\begin{table*}[t]
	\centering
	\setlength{\tabcolsep}{3pt}
	\renewcommand{\arraystretch}{1}
	\resizebox{1\textwidth}{!}{
	\begin{tabular}{lccccccccccccccccccc}
		\toprule \multirow{3}{*}{Method}                                                 & \multicolumn{3}{c}{\textbf{\texttt{CoNeRF Synthetic}}} &                & \multicolumn{3}{c}{\textbf{\texttt{CoNeRF Controllable}}} & \multirow{3}{*}{\textbf{\texttt{GT}}} & \multicolumn{3}{c}{\textbf{\texttt{InterReal \#Medium}}} &                & \multicolumn{3}{c}{\textbf{\texttt{InterReal \#Challenging}}} &            & \multicolumn{3}{c}{\textbf{\texttt{InterReal \#Avg}}} \\
		\cmidrule{2-4} \cmidrule{6-8} \cmidrule{10-12} \cmidrule{14-16} \cmidrule{18-20} & PSNR$\uparrow$                                         & SSIM$\uparrow$ & LPIPS$\downarrow$                                         &                                       & PSNR$\uparrow$                                           & SSIM$\uparrow$ & LPIPS$\downarrow$                                             &            & PSNR$\uparrow$                                       & SSIM$\uparrow$ & LPIPS$\downarrow$ &  & PSNR$\uparrow$  & SSIM$\uparrow$ & LPIPS$\downarrow$ &  & PSNR$\uparrow$  & SSIM$\uparrow$ & LPIPS$\downarrow$ \\
		\midrule                                                                          
		HyperNeRF~\citep{park2021hypernerf}                                              & 25.963                                                 & 0.854          & 0.158                                                     &                                       & 32.520                                                   & 0.981          & 0.169                                                         & \ding{55}  & 25.283                                               & 0.671          & 0.467             &  & 25.261          & 0.713          & 0.517             &  & 25.277          & 0.682          & 0.480             \\
		K-Planes~\citep{kplanes_2023}                                                    & 33.301                                                 & 0.933          & 0.150                                                     &                                       & 31.811                                                   & 0.912          & 0.262                                                         & \ding{55}  & 27.999                                               & 0.813          & 0.177             &  & 26.427          & 0.756          & 0.331             &  & 27.606          & 0.799          & 0.215             \\
		\midrule                                                                          
		CoNeRF~\citep{kania2022conerf}                                                 & 32.394                                                 & 0.972          & 0.139                                                     &                                       & 32.342                                                   & 0.981          & 0.168                                                         & \checkmark & 27.501                                               & 0.745          & 0.367             &  & 26.447          & 0.734          & 0.472             &  & 27.237          & 0.742          & 0.393             \\
		CoGS~\citep{yu2023cogs}                                                          & 33.455                                                 & 0.960          & 0.064                                                     &                                       & 32.601                                                   & \textbf{0.983} & \textbf{0.164}                                                & \checkmark & 30.774                                               & 0.913          & 0.100             &  & ---             & ---            & ---               &  & \textbf{30.774} & 0.913          & 0.100             \\
		LiveScene~\citep{Qu2024LiveSceneLE}                                              & 43.349                                                 & 0.986          & \textbf{0.011}                                            &                                       & 32.782                                                   & 0.932          & 0.186                                                         & \checkmark & 30.815                                               & 0.911          & \textbf{0.066}    &  & 28.436          & 0.846          & 0.185             &  & 30.220          & 0.895          & 0.096             \\
		\midrule MotionGS~\citep{zhu2025motiongs}                                        & 35.057                                                 & 0.981          & 0.052                                                     &                                       & 28.363                                                   & 0.882          & 0.273                                                         & \ding{55}  & 29.193                                               & 0.903          & 0.105             &  & ---             & ---            & ---               &  & 29.193          & 0.903          & 0.105             \\
		\rowhighlight \ours (Ours)                                                       & \textbf{43.939}                                        & \textbf{0.993} & \textbf{0.011}                                            &                                       & \textbf{33.247}                                          & 0.941          & 0.218                                                         & \ding{55}  & \textbf{31.310}                                      & \textbf{0.938} & 0.072             &  & \textbf{29.133} & \textbf{0.899} & \textbf{0.161}    &  & 30.765          & \textbf{0.928} & \textbf{0.094}    \\
		\bottomrule
	\end{tabular}
	}
	\caption{Quantitative results on CoNeRF and InterReal datasets. \ours
	ranks first on CoNeRF synthetic scene and outperforms all competing methods across
	various settings on InterReal datasets.}
    \label{tab:conerf_interreal}
\end{table*}

\begin{table*}[t]
	\centering
	\setlength{\tabcolsep}{3pt}
	\renewcommand{\arraystretch}{1}
	\resizebox{1\textwidth}{!}{
	\begin{tabular}{llccccccccccccccc}
		\toprule \multirow{3}{*}{Method}                & \multirow{3}{*}{\textbf{\texttt{Type}}} & \multirow{3}{*}{\textbf{\texttt{GT}}} & \multicolumn{4}{c}{\textbf{\texttt{\#Easy Sets}}} &                 & \multicolumn{4}{c}{\textbf{\texttt{\#Medium Sets}}} &                   & \multicolumn{4}{c}{\textbf{\texttt{\#Avg (all 20 sets)}}} \\
		\cmidrule{4-7} \cmidrule{9-12} \cmidrule{14-17} &                                         &                                       & M-PSNR$\uparrow$                                  & PSNR$\uparrow$  & SSIM$\uparrow$                                      & LPIPS$\downarrow$ &                                                          & M-PSNR$\uparrow$ & PSNR$\uparrow$  & SSIM$\uparrow$ & LPIPS$\downarrow$ &  & M-PSNR$\uparrow$ & PSNR$\uparrow$  & SSIM$\uparrow$ & LPIPS$\downarrow$ \\
		\midrule                                         
		HyperNeRF~\citep{park2021hypernerf}             & 4D-NeRF                                 & \ding{55}                             & 20.870                                            & 30.708          & 0.908                                               & 0.316             &                                                          & 22.093           & 31.621          & 0.936          & 0.265             &  & 21.679           & 30.748          & 0.917          & 0.299             \\
		K-Planes~\citep{kplanes_2023}                   & 4D-NeRF                                 & \ding{55}                             & 24.211                                            & 32.841          & 0.952                                               & 0.093             &                                                          & 24.312           & 32.548          & 0.954          & 0.100             &  & 24.810           & 32.573          & 0.952          & 0.097             \\
		\midrule CoNeRF~\citep{kania2022conerf}       & Con-NeRF                                & \checkmark                            & 26.561                                            & 32.104          & 0.932                                               & 0.254             &                                                          & 27.716           & 33.256          & 0.951          & 0.207             &  & 27.013           & 32.477          & 0.939          & 0.234             \\
		MK-Planes$^{\star}$                             & Con-NeRF                                & \checkmark                            & 23.509                                            & 31.630          & 0.948                                               & 0.098             &                                                          & 25.860           & 31.880          & 0.951          & 0.104             &  & 24.561           & 31.477          & 0.946          & 0.106             \\
		MK-Planes                                       & Con-NeRF                                & \checkmark                            & 23.872                                            & 31.677          & 0.948                                               & 0.098             &                                                          & 25.217           & 32.165          & 0.952          & 0.099             &  & 24.743           & 31.751          & 0.949          & 0.099             \\
		CoGS~\citep{yu2023cogs}                         & Con-GS                                  & \checkmark                            & 25.208                                            & 32.315          & 0.961                                               & 0.108             &                                                          & 26.332           & 32.447          & 0.965          & 0.086             &  & 26.103           & 32.187          & 0.963          & 0.097             \\
		LiveScene~\citep{Qu2024LiveSceneLE}             & Con-NeRF                                & \checkmark                            & 26.680                                            & \textbf{33.221} & 0.962                                               & \textbf{0.072}    &                                                          & 27.985           & 33.262          & 0.965          & 0.072             &  & 27.310           & 33.158          & 0.962          & 0.072             \\
		\midrule MotionGS~\citep{zhu2025motiongs}       & Flow-GS                                 & \ding{55}                             & 26.306                                            & 31.907          & 0.961                                               & 0.111             &                                                          & 25.391           & 30.904          & 0.969          & 0.083             &  & 25.706           & 31.282          & 0.926          & 0.100             \\
		\rowhighlight \ours (Ours)                      & Flow-GS                                 & \ding{55}                             & \textbf{27.655}                                   & 33.205          & \textbf{0.967}                                      & \textbf{0.072}    &                                                          & \textbf{28.281}  & \textbf{33.922} & \textbf{0.972} & \textbf{0.071}    &  & \textbf{27.838}  & \textbf{33.249} & \textbf{0.969} & \textbf{0.071}    \\
		\bottomrule
	\end{tabular}
	}
	\caption{Quantitative results on \simdata Dataset. \ours surpasses
	prior works on nearly all metrics. “Con-*” indicates Controllable methods, "GT"
	refers to control signals and M-PSNR denotes mask-weighted PSNR for dynamic region.}
    \label{tab:omnisim}
\end{table*}

\subsection{Self-guided Control with Dynamic 3DGS}
\label{sec:self_guided_control} Based on the discussion in~\cref{sec:differential_analysis},
dynamic Gaussian flow constraint \cref{eq:dynamic_gs_flow} provides continuous
Gaussian constraints and, critically, exposes the position of interactive areas,
whose changing topological structures in dynamic scenes are reflected in varying
Gaussian. To overcome the severe dependence on mask annotations in existing
methods, we propose leveraging dynamic Gaussian flow to explore dynamic Gaussians
of interactive objects and extract their trajectories for joint training.

\subsubsection{Dynamic Gaussian clustering and tracking.}
With the formulations in~\cref{eq:dynamic_gs_flow}, we first pretrain a deformable
3DGS $\mathbf{G}^{\prime}$ with a set of camera streams. Then dynamic Gaussian
flow $\mathbf{u}^{\text{GS}}$ from~\cref{eq:dynamic_gs_flow} can be extracted frame-by-frame
and binaried to obtain flow maps. By back-projecting the flow maps to identify
dynamic 3D Gaussians, we highlight Gaussians $\mathcal{D}= \{g_{i}\mid i = 1, 2,
\ldots, Q\}$ with sharp dynamics, as illustrated in~\cref{fig:pipeline}. Next,
we use unsupervised clustering algorithm \textbf{HDBSCAN} to group dynamic Gaussians
into clusters $\mathcal{C}= \{c_{i}\mid i = 1, 2, \ldots, K\}$, where $K$ is the
number of interactive objects. The cluster centers move over time, generating
continuous trajectories $\boldsymbol{\varsigma}(t, k)$, where $k$ indexing which
objects the trajectory belongs to.

\subsubsection{3D Spherical Vector Control.}
Prior works compress control signals into 1D vector. CoNeRF~\citep{kania2022conerf},
takes every control signal as a priori and delegates its encoding to an implicit
MLP that maps to [0, 1]. CoGS~\citep{yu2023cogs}, identifies the start and end
positions of each Gaussian and uses PCA to extract the principal direction of
motion, thereby reducing 3D trajectory to a single 1D vector. Both introduce
fundamental limitations: the 1D vector in CoGS fails to capture complex Gaussian
motions like rotations, while CoNeRF requires the number of controllable regions
and their corresponding signal ranges to be specified in advance, information that
is rarely available in real-world scenarios. We overcome these limitations by
representing the Gaussian states with 3D spherical vectors, which can be directly
obtained from dynamic Gaussian tracking trajectory. This technique eliminates
the requirement of control signals and curve fitting while increasing control
flexibility.

Specifically, in the training stage, we represent the Gaussian dynamics state
using cluster trajectory coordinates
$\mathbf{v}_{c}^{i}= \boldsymbol{\varsigma}(t, k) - \boldsymbol{\varsigma}(0, k)$,
concatenated with Gaussian centers $\mathbf{X}_{i}$. Then, we encode the coordinates
with $\mathbf{E}(\mathbf{v}_{c}^{i}, \mathbf{X}_{i})$ and jointly train the model
$\Theta$ to recover Gaussian dynamics $\left \langle \Delta\mathbf{X}_{i}, \Delta
\mathbf{\Sigma}_{i}\right \rangle$:
\begin{align}
	\boldsymbol{f}_{\Theta}\left(\mathbf{E}(\mathbf{v}_{c}^{i}, \mathbf{X}_{i})\right) \mapsto \left \langle \Delta\mathbf{X}_{i}, \Delta\mathbf{\Sigma}_{i}\right \rangle. \label{eq:training}
\end{align}
After that, we perform splatting rasterization in~\cref{eq:render} with the Gaussian
combining with predicted dynamics. During the control stage, we manually input interactive
3D vector $\mathbf{v}_{c}^{\prime}$, which is mapped to the nearest point in the
original trajectory, to retrive the Gaussian dynamics from the network through $\boldsymbol
{f}_{\Theta}\left(\mathbf{E}(\mathbf{v}_{c}^{\prime}, \mathbf{X}_{i})\right)$.

\subsection{Loss Functions}
\label{sec:loss_functions}
\subsubsection{Loss with dynamic Gaussian flow.}
The expression in~\cref{eq:dynamic_gs_flow} suggests that incorporating optical flow
and camera flow prior to the loss function can improve 3DGS optimization and maintain
dynamic Gaussian smooth transitions between frames. Hence, we propose a dynamic Gaussian
flow loss $\mathcal{L}_{\text{uGS}}$ to optimize the dynamic Gaussian field $\mathbf{G}$
and network $\boldsymbol{\Theta}$ with the following formulation:
\begin{align}
	\mathcal{L}_{\text{uGS}}= \left \| \mathbf{u}- \mathbf{u}^{\text{Cam}}- \sum_{i=1}^{M}T_{i}\alpha_{i}(\boldsymbol{\mu}_{i,t}- \boldsymbol{\mu}_{i,0}) \right \|^{2}, \label{eq:dynamic_gs_flow_loss}
\end{align}
where $\mathbf{u}$ and $\mathbf{u}^{\text{Cam}}$ can be calculated with optical
flow estimator~\citep{2021mmflow} and~\cref{eq:dynamic_gs_flow}, respectively. Dynamic
Gaussians $\mathbf{G}$ and $\boldsymbol{\Theta}$ are optimized via the proposed dynamic
gaussian flow supervision $\mathcal{L}_{\text{uGS}}$ in~\cref{eq:dynamic_gs_flow_loss}
with the fundamental per-frame photometric supervision $\mathcal{L}_{\text{RGB}}$,
and $\mathcal{L}_{\text{D-SSIM}}$. The loss function for \ours optimization can be
formulated as:
\begin{align}
	\mathcal{L}= \lambda \mathcal{L}_{\text{RGB}}+ (1 - \lambda) \mathcal{L}_{\text{D-SSIM}}+ \beta \mathcal{L}_{\text{uGS}}. \label{eq:loss}·
\end{align}
  \section{Experiment}
\label{sec:experiment}

\begin{figure*}[t]
	\begin{center}
		\includegraphics[width=0.95\linewidth]{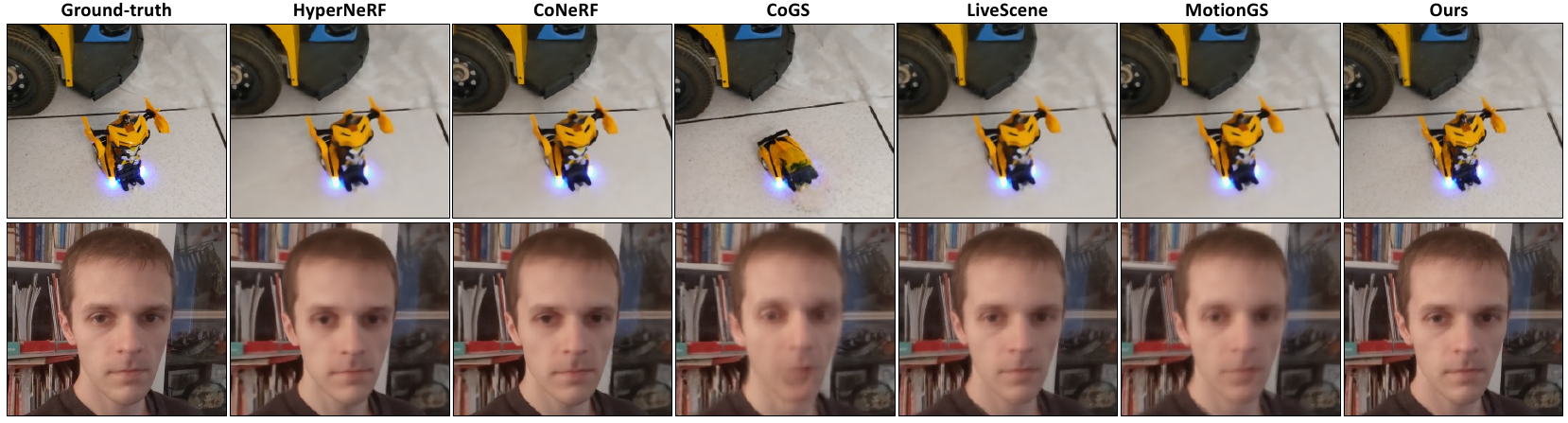}
	\end{center}
	\caption{\textbf{View Synthesis Visualization on CoNeRF Dataset}. In
	comparison with other methods, \ours achieves more realistic and detailed rendering
	quality, whereas other methods suffer from ghosting artifacts.}
	\label{fig:conerf_render}
\end{figure*}

\begin{figure*}[t]
	\begin{center}
		\includegraphics[width=0.95\linewidth]{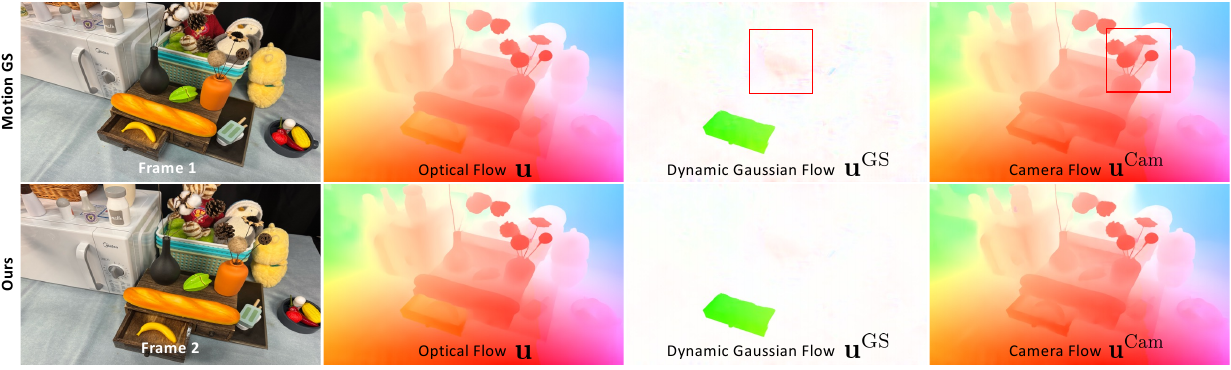}
	\end{center}
	\caption{\textbf{Flow Decoupling Comparison.} \ours (row 2) cleanly separates
	camera egomotion from the microwave’s self-motion, producing artifact-free
	dynamic Gaussian flow.}
	\label{fig:flowmap_compare}
\end{figure*}

\subsection{Experimental Setup}
\label{sec:setup}

\noindent
\textbf{Datasets.} We benchmark~\ours on three publicly-available datasets. We
adopt CoNeRF dataset~\cite{kania2022conerf} for single-object evaluation and ~\simdata
and ~\realdata datasets~\cite{Qu2024LiveSceneLE} for multiple-object setting. A self-captured
toy-kitchen sequence is included for visualization. Complementary novel-view
synthesis are performed on DyNeRF dataset~\cite{li2022neural} (supplementary~\cref{tab:supp_dynerf,fig:supp_dynerf_vis}).
Throughout all experiments the training pipeline remains entirely \textbf{NO
Ground Truth} for control signals.

\noindent
\textbf{Baselines.} \label{sec:baselines} Comparison spans three distinct techniques,
including
3D deformable methods~\cite{kplanes_2023,park2021hypernerf}, controllable scene reconstruction
methods~\cite{kania2022conerf,yu2023cogs,Qu2024LiveSceneLE} and flow-based controllable
method~\cite{zhu2025motiongs}.

\noindent
\textbf{Implementation details.} \label{sec:implementation} \ours is built on
4DGS~\cite{yang2023deformable3dgs}. We use RAFT~\cite{teed2020raft} for optical flow
prediction and perform HDBSCAN clustering for dynamic Gaussian flow with Euclidean
metric. The cluster center is encoded with hash grids and decoded by a MLP. Training
proceeds for 60k steps on a single RTX 4090 with the Adam optimizer at learning rate
$1.6e^{-4}$ in roughly 30 minutes: 30k steps of deformable pre-training followed
by 30k steps of flow training. More details are listed in supplementary~\cref{sec:impl_detail}.

\subsection{Evaluation of Novel View Synthesis}
\label{sec:exp_nvs}

\textbf{Results on CoNeRF Datasets.} The quantitative results of our approach on
the CoNeRF Synthetic and Controllable scenes are presented in~\cref{tab:conerf_interreal}.
Notably, our method surpasses all existing approaches in terms of PSNR, SSIM, and
LPIPS metrics on CoNeRF Synthetic scenes, with a slight advantage over the
second-best method, which benefits from dense labels. Furthermore, on CoNeRF Controllable
scenes, our method attains the highest PSNR of 33.247, while demonstrating comparable
SSIM and LPIPS scores to the SOTA methods. These results underscore the success
of the guidance-free paradigm. \cref{fig:conerf_render} visualizes the rendering
result of our method on the CoNeRF dataset. Our method handles the controllable
objects well and retains the details of the moving area, demonstrating its effectiveness
in modeling interactive scenes.

\noindent
\textbf{Metric on LiveScene Dataset.} As reported in \cref{tab:omnisim,tab:conerf_interreal},
\ours leads across both \simdata and \realdata while remaining fully annotation-free.
On \simdata, it achieves the highest scores on \texttt{\#medium} subset,
surpassing sparse-label baselines~\cite{zhu2025motiongs} by nearly 2dB in PSNR. Although
PSNR is slightly inferior to the dense-labele LiveScene on the \texttt{\#easy}
subset, its advantage is decisive whenever manual labels are unavailable. M-PSNR
metric further confirms superior reconstruction qaulity of dynamic regions. On \realdata,
CoGS and MotionGS underperform on \texttt{\#medium} and collapses on the \texttt{\#challenging}
subset, where prolonged trajectories and dense interaction expose the limits of prior
controllable or flow-based methods. \ours not only converges robustly but also
posts the best \texttt{\#challenging} results and the top \texttt{\#medium} PSNR
and SSIM, demonstrating robust fidelity and stability in large-scale, real-world
interactive scenarios with incomplete supervision. Visualization comparisions can
be found on supplementary~\cref{fig:supp_render_sim}.

\begin{figure}[t]
	\begin{center}
		\includegraphics[width=0.91\linewidth]{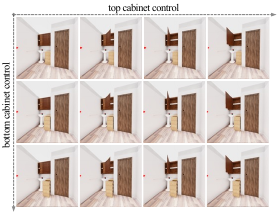}
	\end{center}
	\caption{\textbf{Individual Object Control}. Our method supports per-object
	manipulation, enabling the synthesis of previously unseen views for each scene
	without retraining.}
	\label{fig:dual_control}
\end{figure}

\noindent
\textbf{Individual Object Control Visualiztion.} \cref{fig:dual_control} presents
a example to demonstrate case of per-object control. During manipulation, each object
is assigned an independent 3D spherical vector which controls its instantaneous motion.
This disentanglement removes cross-object constraints, and allows the model to compose
attribute combinations absent from the training set. The example demonstrates a
sequence where two cabinets always open or close together. By independently setting
their control vectors, we generate a configuration in which the top cabinet is
open while the bottom cabinet remains closed (top-right), confirming that the model
can extrapolate novel scene arrangements with both diversity and fidelity.

\begin{figure*}[t]
	\begin{center}
		\includegraphics[width=0.93\linewidth]{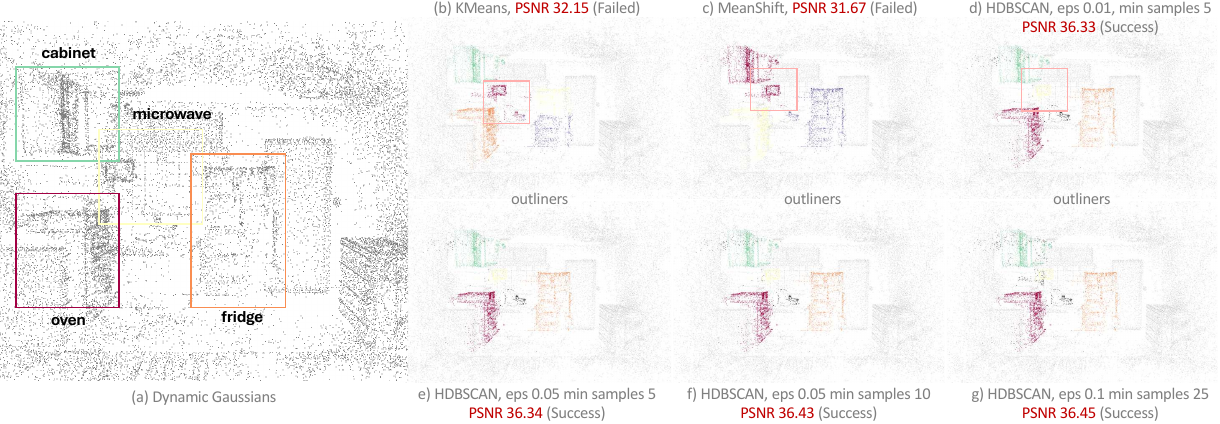}
	\end{center}
	\caption{Ablation of clustering results among KMeans, MeanShift and HDBSCAN on
	\#seq001 of \simdata.}
	\label{fig:cluster_sim}
\end{figure*}

\noindent
\textbf{Flow Decouple Visualization.} \cref{fig:flowmap_compare} contrasts the flow-decomposition
quality of \ours and MotionGS on real-life toy-kitchen scene, in which an automatically
opening drawer and a moving camera jointly generate complex optical flow.
Although both methods estimate the optical flow accurately, MotionGS decouples camera
flow that is both noisy and partially aliased (top-right), thus the residual
dynamic gaussian flow inherits substantial artifacts. In contrast, \ours cleanly
disentangles the camera-induced flow from the drawer's independent motion, yielding
a per-object component that is significantly cleaner. This precise separation
supplies downstream constraints with more reliable guidance, thus improving the render
fidelity.

\subsection{Ablation and analysis}


We conduct ablation studies to examine the contribution of two components in \ours.
Following previous work~\cite{Qu2024LiveSceneLE}, we select three representative
subsets from the \simdata dataset: \texttt{\#seq001}, \texttt{\#seq004}, and
\texttt{\#seq0015} and a self-captured toy-kitchen dataset. ~\cref{tab:ablation}
shows the results of each ablation experiment.

\begin{table}[h]
	\centering
	\setlength{\tabcolsep}{8pt}
	\renewcommand{\arraystretch}{1}
	\resizebox{1\columnwidth}{!}{
	\begin{tabular}{llccc}
		\toprule \multicolumn{2}{c}{\textbf{\texttt{Setting}}} & PSNR$\uparrow$                        & SSIM$\uparrow$ & LPIPS$\downarrow$ \\
		\midrule \multirow{4}{*}{Sim}                          & \ours                                 & \textbf{35.31} & \textbf{0.975}   & \textbf{0.062} \\
		                                                       & \#1.HDBSCAN $\rightarrow$ KMeans      & 31.33          & 0.966            & 0.065          \\
		                                                       & \#2.HDBSCAN $\rightarrow$ MeanShift   & 30.95          & 0.959            & 0.068          \\
		                                                       & \#3.3D Vector $\rightarrow$ 1D Vector & 33.22          & 0.969            & 0.064          \\
		\midrule \multirow{4}{*}{Real}                         & \ours                                 & \textbf{32.45} & \textbf{0.951}   & 0.092          \\
		                                                       & \#4.HDBSCAN $\rightarrow$ KMeans      & 32.33          & 0.949            & \textbf{0.091} \\
		                                                       & \#5.HDBSCAN $\rightarrow$ MeanShift   & 31.86          & 0.932            & 0.100          \\
		                                                       & \#6.3D Vector $\rightarrow$ 1D Vector & 30.33          & 0.918            & 0.107          \\
		\bottomrule
	\end{tabular}
	} 
	\caption{\textbf{Ablation Study.} Ablations on two components of our proposed
    \label{tab:ablation}
	method.}
\end{table}

\begin{figure}[t]
	\begin{center}
		\includegraphics[width=1\linewidth]{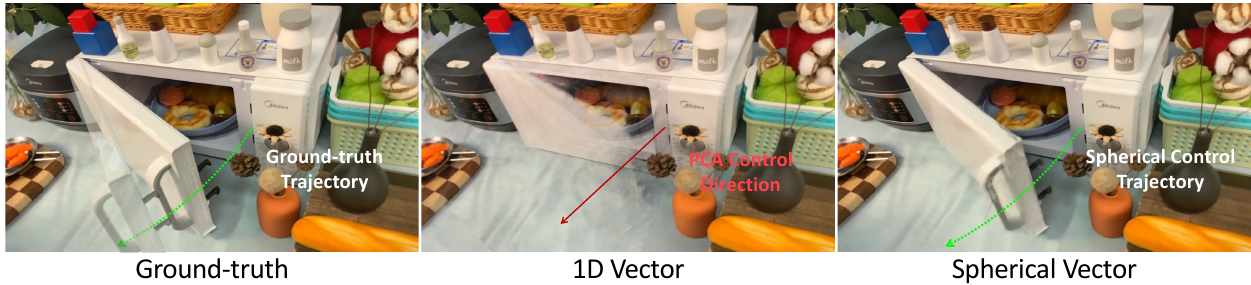}
	\end{center}
	\caption{\textbf{Ablation of 3D spherical vector.} 1D vector PCA could not match
	arc trajectory, while 3D spherical vectors recover fine structure and motion.}
	\label{fig:3d_1d_control}
\end{figure}

\noindent
\textbf{Effectiveness of 3D Vector Control.} To validate the necessity of 3D vector,
we conduct ablation using directly 1D vector adopted by CoGS while keeping all
other settings identical. As shown in the~\cref{tab:ablation} (\#3, \#6), this change
degrades rendering quality since PCA only approximates the dominant direction,
leaving detailed trajectories misaligned, shown in the middle of ~\cref{fig:3d_1d_control}.
Consequently, the model reconstructs coarse structures in the control stage. In contrast,
3D vector provides per-Gaussian clusters, fine-grained control (right); the
explicit motion cues tightly constrain the Gaussian flow, ensuring consistent motion
guidance between training and controlling.

\noindent
\textbf{Effectiveness of HDBSCAN Clustering.} Clustering is essential in the control
stage as the number of controllable objects is not a prior of our approach.Compared
with widely used clustering methods like KMeans, HDBSCAN is more robust to noise
with outliner handling and more flexible without predefined cluster numbers. Besides,
MeanShift may converge to local optima depending on the cluster landscape and
initial window locations. ~\cref{fig:cluster_sim} illustrates the remarkable
stability and accuracy of HDBSCAN. (d)-(g) show that, HDBSCAN delineates object
geometry cleanly and isolates outliers, whereas K-Means introduces a large number
of noisy points (b) and Meanshift yields an inappropriate cluster cardinality (c).
For real life scene clustering visualizations, please refer to the supplementary
materials.

  \section{Conclusion}
\label{sec:conclusion}
In this work, we establish a mathematical link among optical flow, camera flow, and dynamic Gaussian flow with differential analysis, yielding an annotation-free Gaussian-splatting pipeline for controllable view synthesis. Flow-based constraints refine optimization, ensuring smooth motion and high fidelity while highlighting interactable Gaussians without manual labels. After obtaining each individual object in the scene, a 3D spherical vector further encodes object state, eliminating explicit trajectory computation. Extensive experiments demonstrate our superior performance in both view synthesis and scene controlling, enabling more accurate and efficient modeling of articulated objects.

\noindent
\textbf{Limitations:} \ours relies on optical flow estimator and may compromise view synthesis or control robustness under lighting variation. Failure cases are shown in the supplementary materials~\cref{fig:supp_flow_failure}.


  {\small

\boldparagraph{Acknowledgements.}
This work is supported by the Shanghai AI Laboratory and the National Natural Science Foundation of China (624B2044).
}
  \bibliography{aaai2026}
  \clearpage
\setcounter{page}{1}

\begin{abstract}
    \noindent
    This supplementary material accompanies the main paper by providing more details
    for reproducibility as well as additional evaluations and qualitative results
    to verify the effectiveness and robustness of \ours:\\
    \noindent
    $\triangleright$ \textbf{\cref{sec:supp_differential_analysis}}: Dynamic Gaussian
    Flow Derivative Proof. \\
    \noindent
    $\triangleright$ \textbf{\cref{sec:impl_detail}}: Additional implementation
    details. \\
    \noindent
    $\triangleright$ \textbf{\cref{sec:add_exp}}: Additional experimental results,
    including more detailed view synthesis quality comparison, clustering visualization,
    illustrations of gaussian flow map and failure cases of our method.
\end{abstract}

\section{Detailed Dynamic Gaussian Flow Analysis}
\label{sec:supp_differential_analysis} Our insight is that dynamic Gaussian flow
under instantaneous motion can be analytically decoupled from optical flow and
camera motion via differential analysis with alpha composition. Considering a dynamic
scene with interactive objects as shown in~\cref{fig:dynamic_gaussian}, the
camera and 3D Gaussians hold separate velocities in consecutive frames $0$ and
$t$. Assuming a dynamic 3D Gaussian $G_{i}$ with velocity $\boldsymbol{v}^{\text{GS}}$,
it is projected as image measurement $g_{i}$ under the constant camera
instantaneous motion by translation velocity $\boldsymbol{v}$ and rotational
velocity $\boldsymbol{\omega}$. The optical flow $\mathbf{u}$ induced by $(\boldsymbol
{v}, \boldsymbol{\omega})$ of a pixel $\mathbf{m}= (x,y)^{\top}$ can be obtained
by \emph{Lemma 1}:

\boldparagraph{Lemma 1:} \textit{Dynamic Gaussian flow $\mathbf{u}^{\text{GS}}$
under instantaneous motion can be derived from optical flow $\mathbf{u}$ and
camera flow $\mathbf{u}^{\text{Cam}}$ with the following transform~\cref{eq:supp_gaussian_flow_analysis}.}
\begin{equation}
    \small
    \begin{aligned}
        \label{eq:supp_gaussian_flow_analysis} & \mathbf{u}= \mathbf{u}^{\text{Cam}}+ \mathbf{u}^{\text{GS}}+ \mathbf{\Delta}, \quad \mathbf{u}^{\text{Cam}}= \frac{\mathbf{A}\boldsymbol{v}}{Z}+ \mathbf{B}\boldsymbol{\omega},                                               \\
        \\
                                               & \mathbf{u}^{\text{GS}}= \mathbf{A}\sum_{i=1}^{M}T_{i}\alpha_{i}\frac{\boldsymbol{v}^\text{GS}}{Z_i}, \mathbf{\Delta}= \mathbf{A}\sum_{i=1}^{M}T_{i}\alpha_{i}\boldsymbol{v}(\frac{1}{Z_i}- \frac{1}{Z}),                      \\
                                               & \mathbf{A}= \begin{bmatrix}-f_{x}&0&x - c_{x}\\ 0&-f_{y}&y - c_{y}\end{bmatrix},                                                                                                                                              \\
                                               & \mathbf{B}= \begin{bmatrix}\frac{(x - c_x)(y - c_y)}{f_y}&- f_{x}- \frac{(x - c_x)^2}{f_x}&\frac{(y - c_y) f_x}{f_y}\\ f_{y}+ \frac{(y - c_y)^2}{f_y}&-\frac{(x - c_x)(y - c_y)}{f_x}&-\frac{(x - c_x)f_y}{f_x}\end{bmatrix}.
    \end{aligned}
\end{equation}
where $f_{x}, f_{y}, c_{x}, c_{y}$ are camera intrinsics, $M$ denotes the number
of Gaussian projections sorted with Gaussian depth $Z_{i}$ intersecting the pixel
$\mathbf{m}$. Flow residual term $\mathbf{\Delta}$ are preserved to guarantee
accuracy, even when they approach zero after refined optimization.

\textit{Proof.} We first derive the formula for 3D Gaussians derivative induced
by camera rotation $\mathbf{R}(t)$, translation $\mathbf{T}(t)$, and Gaussian
translation $\mathbf{T}^{\text{GS}}(t)$, which transform the 3D Gaussian $G_{i}$
under constant instantaneous-motion as time $t$ increasing. The equation
transforming Gaussian $G_{i}$ from time $t$ to $0$ can be formulated as:
\begin{align}
    \mathbf{X}_{i}(0) - \mathbf{T}_{i}^{\text{GS}}(t) = \mathbf{R}(t)\mathbf{X}_{i}(t) + \mathbf{T}(t), \label{eq:supp_gs_transform}
\end{align}
By derivative in both sides, we reformulate the Gaussian transform in~\cref{eq:supp_gs_transform}
as:
\begin{align}
    \label{eq:supp_point_derivative_transform}-\dot{\mathbf{T}}_{i}^{\text{GS}}(t) & = \dot{\mathbf{R}}(t)\mathbf{X}_{i}(t) + \mathbf{R}(t)\dot{\mathbf{X}_i}(t) + \dot{\mathbf{T}}(t), \\
    \dot{\mathbf{X}_i}(t)                                                          & = -\mathbf{R}^{\top}(t)\dot{\mathbf{R}}(t)\mathbf{X}_{i}(t) \notag                                 \\
                                                                                   & \quad - \mathbf{R}^{\top}(t)\dot{\mathbf{T}}(t)                                                    \\
                                                                                   & \quad - \mathbf{R}^{\top}(t)\dot{\mathbf{T}}_{i}^{\text{GS}}(t). \notag
\end{align}
According to Possion's equation~\citep{poissonequation,heeger1992subspace}, the rotation
and translation velocities can be defined with $\mathbf{R}^{\top}(t)\dot{\mathbf{R}}
(t) = [\boldsymbol{\omega}]_{\times}$,\quad$\mathbf{R}^{\top}(t)\dot{\mathbf{T}}(
t) = \boldsymbol{v}$
and
$\mathbf{R}^{\top}(t)\dot{\mathbf{T}^\text{GS}}(t) = \boldsymbol{v}^{\text{GS}}$.
By substituting the above equations into~\cref{eq:supp_point_derivative_transform}
and omitting the time notation, we obtain the simplicity results:
\begin{align}
    \dot{\mathbf{X}}_{i} & = -[\boldsymbol{\omega}]_{\times}\mathbf{X}_{i}- \boldsymbol{v}- \boldsymbol{v}^{\text{GS}}, \label{eq:dX_dt}
\end{align}
where $\boldsymbol{v}^{\text{GS}}$ presents the velocity of the dynamic 3D Gaussian
$G_{i}$. Then, the camera projection model with respect to $\mathbf{X}_{i}$ is:
\begin{equation}
    \begin{aligned}
        Z_{i}[\mathcal{\mu}_{i}; 1] = \mathbf{KX}_{i}.
    \end{aligned}
\end{equation}
In order to derive the dynamic Gaussian flow $\mathbf{u}_{i}^{\text{GS}}$ in the
2D image plane, we derivative on both sides and obtain the differential of the projected
image coordinates, namely the optical flow, in relation to the projection parameters:
\begin{equation}
    \begin{aligned}
        \label{eq:supp_camera_projection_derivative}\mathbf{u}_{i}^{\text{GS}}= \begin{bmatrix}\frac{f_{x}}{Z} & 0 & -\frac{f_{x} X}{Z^{2}} \\ 0 & \frac{f_{y}}{Z} & -\frac{f_{y} Y}{Z^{2}}\end{bmatrix} \dot{\mathbf{X}}_{i}.
    \end{aligned}
\end{equation}
By substituting the above equations~\cref{eq:supp_camera_projection_derivative}
into~\cref{eq:dX_dt}, we obtain the dynamic Gaussian flow decomposition $\mathbf{u}
_{i}^{\text{GS}}$ in individual Gaussian $G_{i}$ as:
\begin{equation}
    \begin{aligned}
        \mathbf{u}_{i} & = \frac{\mathbf{A}\boldsymbol{v}}{Z_i}+ \mathbf{B}\boldsymbol{\omega}+ \frac{\mathbf{A}\boldsymbol{v}^\text{GS}}{Z_i},                                                                                                     \\
                       & = \left( \frac{\mathbf{A}\boldsymbol{v}}{Z}+ \mathbf{B}\boldsymbol{\omega}\right) + \frac{\mathbf{A}\boldsymbol{v}^\text{GS}}{Z_i}+ \left( \frac{\mathbf{A}\boldsymbol{v}}{Z_i}- \frac{\mathbf{A}\boldsymbol{v}}{Z}\right)
    \end{aligned}
\end{equation}
With alpha composition, we weight the flow with $w_{i}=\frac{T_{i}\alpha_{i}}{\Sigma_{i}T_{i}\alpha_{i}}$
in both sides and proof the mathematical relation described in~\cref{eq:supp_gaussian_flow_analysis}.
\hfill $\square$

\textit{Proof.} Assuming the Gaussian to be isotropic~\cite{gao2024gaussianflow},
with covariance matrix $\mathbf{B}_{i,t}\mathbf{B}_{i,t}^{\top}=\mathbf{RS}\mathbf{S}
^{\top}\mathbf{R}^{\top}=\sigma^{2}\mathbf{I}$. With a constant instantaneous-motion
model, the tiny varation of scaling factor $\sigma$ of each Gaussian can be simply
ignored, and $\mathbf{B}_{i,t}\mathbf{B}^{-1}_{i,0}\approx \mathbf{I}$. Therefore,
the projection flow of a dynamic Gaussian $G_{i}$ varying from $0$ to $t$ can be
formulated as
$\mathbf{\tilde{u}}_{i}^{\text{GS}}= \boldsymbol{\mu}_{i,t}- \boldsymbol{\mu}_{i,0}$.
The difference between two Gaussian-distributed variables $\mathbf{m}_{i,0}$ and
$\mathbf{m}_{i,t}$ can be expressed as:
\begin{equation}
    \begin{aligned}
        \label{eq:isotropic_gs}\mathbf{\tilde{u}}_{i}^{\text{GS}} & = \mathbf{x}_{i,t}- \mathbf{x}_{i,0}                                                                                     \\
                                                                  & = \mathbf{B}_{i,t}\mathbf{B}^{-1}_{i,0}(\mathbf{x}_{0}- \boldsymbol{\mu}_{i,t}) + \boldsymbol{\mu}_{i,t}- \mathbf{x}_{0} \\
                                                                  & = \boldsymbol{\mu}_{i,t}- \boldsymbol{\mu}_{i,0}.
    \end{aligned}
\end{equation}

By weighting the flow on both side, and substituting the flow into~\cref{eq:gaussian_flow_analysis},
we obtain the relation among the optical flow, camera flow, and dynamic Gaussian
flow. Note that the isotropic Gaussian assumption helps to reduce computational
complexity and enhance optimization stability. It is a common practice in many works~\citep{gao2024gaussianflow,ling2024align,keetha2024splatam}.
Nevertheless, it is still flexible to extend to anisotropic in practice with~\cref{eq:isotropic_gs}.

\section{Additional implementation details}
\label{sec:impl_detail}

\boldparagraph{Implementation Details.} \ours is built on 4DGS~\cite{yang2023deformable3dgs}.
We use RAFT~\cite{teed2020raft} for optical flow prediction and perform HDBSCAN clustering
from dynamic Gaussian flow with Euclidean metric, $\epsilon = 0.05$, minimal samples
= 5 and min cluster size = 400. The cluster center corresponding to each Gaussian
is encoded with hash grids and decoded with an 8-layer MLP with 256 neurons. The
model is trained on an NVIDIA GeForce RTX 4090 GPU for 60k steps, using Adam optimizer
with learning rate $1.6e^{-4}$ and batch size 1. The coarse-to-fine training process
lasts 30 minutes and is divided into 3 stages, including 500 steps of canonical
warmup, 30k steps 4d deformable training, and 30k steps of full training. For all
experiments, we set loss weights of $\mathcal{L}_{\text{RGB}}$, $\mathcal{L}_{\text{D-SSIM}}$,
and $\mathcal{L}_{\text{uGS}}$ as $\lambda = 0.8$, $(1 - \lambda) = 0.2$, and
$\beta = 0.5$, respectively.

\boldparagraph{Dynamic Gaussian Clustering.} Gaussian clustering would impact
the control ability of the model, which in turn is directly influenced by the quality
of the back-projected flow map. We configure the frame interval to be 1 and
establish correspondences between the optical flows of adjacent frames. By leveraging~\cref{eq:supp_gaussian_flow_analysis},
we compute the Gaussian interaction flow. Next, by randomly sampling 5\% of the interaction
flow map as keyframes, we perform back-projection and apply HDBSCAN clustering
to obtain dynamic Gaussians. Small keyframe ratios lead to incomplete clustering,
while a 5\% ratio is sufficient for achieving better clustering results.
Conversely, higher ratios result in noisy clustering, which hinders subsequent
control.


\boldparagraph{Algorithm Implementation.} Algorithm~\ref{algorithm:supp_alg}
provided detailed implementation pseudo code of \ours, including the deformable 3D
Gaussian pre-training, dynamic Gaussian flow decouple, HDBSCAN clustering, and
Self-guide control with dynamic 3D Gaussian.
\begin{algorithm}
    [ht] \SetKwInOut{Input}{Input} \SetKwInOut{Output}{Output} \Input{Set camera stream \{$\mathbf{P}(t)$, $\mathbf{I}(t)$\} and initialize 3D Gaussians $\mathbf{G}^{0}$.}
    \Output{Controllable 3D Gaussians $\mathbf{G}^{\ast}$ with Network $\Theta^{\ast}$.}
    $\rhd$ pre-train a deformable 3DGS $\mathbf{G}^{\prime}$\;
    $\bigtriangledown$ Dynamic Gaussian Flow Decouple\; \For{Each continuous camera views $\mathbf{P}(0), \mathbf{P}(t)$}
    { Estimate optical flow $\mathbf{u}$ and caculate camera flow $\mathbf{u}^{\text{Cam}}$ using~\cref{eq:supp_gaussian_flow_analysis}\; Calculate dynamic gaussian flow $\mathbf{u}^{\text{GS}}$ using~\cref{eq:dynamic_gs_flow}\; Back project binarized dynamic Gaussian flow $\textbf{bin}(\mathbf{u}^{\text{GS}})$ to 3DGS: $g_{i}\to \mathcal{D}$\; }
    $\rhd$ $\textbf{HDBSCAN}$ clustering and caculate trajactory
    $\boldsymbol{\varsigma}(t, k)$\; $\bigtriangledown$ Self-guided Control with
    Dynamic 3DGS\; \While{(not reach max iteration) and (not satisfy stopping criteria)}
    { \For{Each continuous pair $<\mathbf{P}(t), \mathbf{I}(t)>$} { Encode coordinates $\mathbf{v}_{c}^{i}= \boldsymbol{\varsigma}(t, k) - \boldsymbol{\varsigma}(0, k)$ with hash grid: $\textbf{E}(\mathbf{v}_{c}^{i})$\; Forward pass and rasterize with $\mathbf{G}^{\ast}$ and $\textbf{E}(\boldsymbol{\varsigma})$: $\mathbf{I}, \mathbf{u}^{\text{GS}}$ = $\Theta(\mathbf{G}^{\ast}, \textbf{E}(\boldsymbol{\varsigma}))$\; Calculate loss $\mathcal{L}_{\text{uGS}}$, $\mathcal{L}_{\text{RGB}}$, $\mathcal{L}_{\text{D-SSIM}}$ using~\cref{eq:dynamic_gs_flow} and optimize with Gradient Descent\; Update $\boldsymbol{\Theta^\ast}$ and $\mathbf{G}^{\ast}$\; } }
    $\bigtriangledown$ Controlling with \ours\; \For{Each control camera view and 3d vector $\mathbf{v}_{c}^{\prime}$}
    { Back-project to query Gaussian $G_{i}$ \; Perform hash encoding: $\textbf{E}(\mathbf{v}_{c}^{\prime})$\; Forward pass $\Theta^{\ast}$ and rasterize with $\boldsymbol{f}_{\Theta^\ast}\left(\mathbf{X}_{i}, \mathbf{v}_{c}^{\prime}\right)$ }
    \caption{Controllable 3D Gaussian Splats with Flow Derivatives}
    \label{algorithm:supp_alg}
\end{algorithm}

\section{Additional Experimental Results}
\label{sec:add_exp}

\subsection{Evaluation of efficiency}
\label{sec:exp_efficiency} To better demonstrate the advantages of \ours, we
picked \#seq002 from the \simdata for statistical modeling of the number of parameters,
running memory and rendering speed. ~\cref{tab:efficiency} describes that our
method achieves a rendering speed of 123.88 FPS, which is significantly faster than
NeRF based methods, while maintaining a relatively low memory footprint of 5.43
GB. The number of parameters in \ours is 49.84 MB, which is smaller than 1/4 the
size of CoGS. These results shows that \ours is not only efficient in terms of memory
usage and rendering speed but also has a smaller model size compared to existing
methods.

\begin{table}[t]
    \centering
    \setlength{\tabcolsep}{2pt}
    \renewcommand{\arraystretch}{1}
    \caption{\textbf{Model performance across size and speed}. We show the
    comparison of model performance in terms of number of parameters, rendering speed,
    and runtime memory.}
    \resizebox{1\columnwidth}{!}{
    \begin{tabular}{lccccc}
        \toprule Method                          & Batch size & Ray samples & FPS    & Parameters (MB) & Memory (GB) \\
        \midrule CoNeRF~\cite{kania2022conerf} & 1024       & 256         & 0.22   & 149.58          & 71.93       \\
        MK-Planes~\cite{kplanes_2023}            & 4096       & 48          & 2.07   & 154.19          & 12.48       \\
        MK-Planes*~\cite{kplanes_2023}           & 4096       & 48          & 0.61   & 152.35          & 11.90       \\
        LiveScene~\cite{Qu2024LiveSceneLE}       & 4096       & 48          & 0.62   & 144.80          & 8.24        \\
        \midrule CoGS~\cite{yu2023cogs}          & 1          & -           & 215.93 & 189.70          & 25.50       \\
        MotionGS~\cite{zhu2025motiongs}          & 1          & -           & 105.30 & 404.77          & 60.34       \\
        \ours (Ours)                             & 1          & -           & 123.88 & 49.84           & 5.43        \\
        \bottomrule
    \end{tabular}
    } \label{tab:efficiency}
\end{table}

\begin{table*}
    [ht]
    \centering
    \setlength{\tabcolsep}{8pt}
    \renewcommand{\arraystretch}{1.0}
    \caption{\textbf{Detailed Quantitative Results on \simdata Dataset}. \ours
    outperforms prior works on most metrics, especially the \#easy and \#medium subsets.}
    \label{tab:supp_ominisim} \resizebox{2.0\columnwidth}{!}{
    \begin{tabular}{l|lccccccccc|c}
        \hline
        Dataset                     & Metric & NeRF   & Instant-NGP & HyperNeRF & CoNeRF & K-Planes & MK-Planes & MK-Planes$^{\ast}$ & LiveScene & CoGS   & \ours  \\
        \hline
        seq001\_Rs\_int             & psnr   & 25.941 & 25.768      & NaN       & 34.035 & 33.136   & 32.169    & 32.092             & 34.784    & 32.211 & 36.335 \\
        seq001\_Rs\_int             & ssim   & 0.931  & 0.933       & NaN       & 0.957  & 0.953    & 0.946     & 0.946              & 0.974     & 0.968  & 0.980  \\
        seq001\_Rs\_int             & lpips  & 0.118  & 0.113       & NaN       & 0.135  & 0.093    & 0.110     & 0.110              & 0.048     & 0.068  & 0.046  \\
        seq002\_Rs\_int             & psnr   & 28.616 & 28.660      & NaN       & 34.286 & 34.765   & 36.532    & 34.580             & 35.190    & 34.497 & 34.979 \\
        seq002\_Rs\_int             & ssim   & 0.950  & 0.946       & NaN       & 0.951  & 0.967    & 0.976     & 0.968              & 0.969     & 0.979  & 0.976  \\
        seq002\_Rs\_int             & lpips  & 0.096  & 0.112       & NaN       & 0.217  & 0.074    & 0.036     & 0.074              & 0.070     & 0.051  & 0.060  \\
        seq003\_Ihlen\_1\_int       & psnr   & 26.720 & 28.255      & 33.551    & 34.700 & 35.217   & 34.758    & 34.753             & 35.323    & 36.816 & 36.094 \\
        seq003\_Ihlen\_1\_int       & ssim   & 0.940  & 0.944       & 0.946     & 0.953  & 0.964    & 0.966     & 0.966              & 0.966     & 0.980  & 0.974  \\
        seq003\_Ihlen\_1\_int       & lpips  & 0.120  & 0.121       & 0.268     & 0.244  & 0.097    & 0.087     & 0.090              & 0.094     & 0.077  & 0.077  \\
        seq004\_Ihlen\_1\_int       & psnr   & 30.847 & 31.800      & 31.115    & 32.684 & 36.157   & 34.863    & 35.000             & 36.712    & 31.055 & 35.700 \\
        seq004\_Ihlen\_1\_int       & ssim   & 0.927  & 0.942       & 0.878     & 0.888  & 0.955    & 0.919     & 0.926              & 0.962     & 0.915  & 0.965  \\
        seq004\_Ihlen\_1\_int       & lpips  & 0.104  & 0.102       & 0.389     & 0.366  & 0.085    & 0.145     & 0.135              & 0.072     & 0.209  & 0.086  \\
        seq005\_Beechwood\_0\_int   & psnr   & 27.183 & 27.295      & 30.699    & 32.549 & 31.944   & 33.195    & 33.098             & 33.623    & 33.664 & 33.778 \\
        seq005\_Beechwood\_0\_int   & ssim   & 0.930  & 0.937       & 0.906     & 0.927  & 0.944    & 0.961     & 0.959              & 0.962     & 0.978  & 0.973  \\
        seq005\_Beechwood\_0\_int   & lpips  & 0.127  & 0.112       & 0.291     & 0.245  & 0.105    & 0.076     & 0.080              & 0.072     & 0.058  & 0.063  \\
        seq006\_Beechwood\_0\_int   & psnr   & 27.988 & 28.150      & 29.513    & 30.058 & 31.861   & 31.541    & 31.521             & 32.206    & 31.272 & 32.067 \\
        seq006\_Beechwood\_0\_int   & ssim   & 0.938  & 0.938       & 0.907     & 0.917  & 0.951    & 0.951     & 0.951              & 0.959     & 0.974  & 0.971  \\
        seq006\_Beechwood\_0\_int   & lpips  & 0.103  & 0.119       & 0.314     & 0.283  & 0.097    & 0.095     & 0.096              & 0.077     & 0.059  & 0.058  \\
        seq007\_Beechwood\_0\_int   & psnr   & 23.201 & 22.902      & 31.259    & 33.451 & 30.979   & 30.136    & 30.089             & 30.360    & 27.367 & 33.748 \\
        seq007\_Beechwood\_0\_int   & ssim   & 0.885  & 0.886       & 0.913     & 0.935  & 0.938    & 0.942     & 0.942              & 0.946     & 0.893  & 0.969  \\
        seq007\_Beechwood\_0\_int   & lpips  & 0.220  & 0.219       & 0.289     & 0.229  & 0.140    & 0.120     & 0.121              & 0.107     & 0.219  & 0.084  \\
        seq008\_Benevolence\_1\_int & psnr   & 25.750 & 25.574      & 32.691    & 34.319 & 31.914   & 30.926    & 30.916             & 33.393    & 33.795 & 33.855 \\
        seq008\_Benevolence\_1\_int & ssim   & 0.943  & 0.940       & 0.945     & 0.960  & 0.948    & 0.941     & 0.941              & 0.970     & 0.980  & 0.975  \\
        seq008\_Benevolence\_1\_int & lpips  & 0.113  & 0.123       & 0.229     & 0.185  & 0.107    & 0.118     & 0.116              & 0.067     & 0.072  & 0.068  \\
        seq009\_Benevolence\_1\_int & psnr   & 24.326 & 24.386      & 29.596    & 31.225 & 32.836   & 31.500    & 31.471             & 32.030    & 33.205 & 31.960 \\
        seq009\_Benevolence\_1\_int & ssim   & 0.921  & 0.922       & 0.897     & 0.932  & 0.956    & 0.954     & 0.953              & 0.962     & 0.975  & 0.959  \\
        seq009\_Benevolence\_1\_int & lpips  & 0.124  & 0.128       & 0.327     & 0.248  & 0.090    & 0.088     & 0.090              & 0.071     & 0.074  & 0.089  \\
        seq010\_Merom\_1\_int       & psnr   & 22.927 & 22.765      & 28.985    & 31.092 & 30.120   & 29.461    & 29.396             & 30.029    & 30.254 & 30.622 \\
        seq010\_Merom\_1\_int       & ssim   & 0.917  & 0.925       & 0.939     & 0.957  & 0.960    & 0.960     & 0.959              & 0.966     & 0.974  & 0.971  \\
        seq010\_Merom\_1\_int       & lpips  & 0.173  & 0.158       & 0.275     & 0.233  & 0.093    & 0.087     & 0.088              & 0.074     & 0.065  & 0.080  \\
        seq011\_Merom\_1\_int       & psnr   & 26.732 & 27.077      & NaN       & 30.483 & 33.394   & 32.951    & 32.910             & 33.426    & 31.767 & 33.014 \\
        seq011\_Merom\_1\_int       & ssim   & 0.932  & 0.933       & NaN       & 0.932  & 0.959    & 0.959     & 0.959              & 0.960     & 0.968  & 0.966  \\
        seq011\_Merom\_1\_int       & lpips  & 0.112  & 0.117       & NaN       & 0.246  & 0.074    & 0.073     & 0.072              & 0.068     & 0.091  & 0.079  \\
        seq012\_Pomaria\_1\_int     & psnr   & 26.856 & 27.074      & NaN       & 33.065 & 35.185   & 32.248    & 32.209             & 33.367    & 37.284 & 34.104 \\
        seq012\_Pomaria\_1\_int     & ssim   & 0.936  & 0.943       & NaN       & 0.954  & 0.972    & 0.966     & 0.966              & 0.969     & 0.985  & 0.972  \\
        seq012\_Pomaria\_1\_int     & lpips  & 0.138  & 0.126       & NaN       & 0.199  & 0.059    & 0.075     & 0.075              & 0.061     & 0.047  & 0.067  \\
        seq013\_Pomaria\_1\_int     & psnr   & 25.277 & 24.018      & NaN       & 33.682 & 30.860   & 30.390    & 30.299             & 33.592    & 32.868 & 32.730 \\
        seq013\_Pomaria\_1\_int     & ssim   & 0.925  & 0.930       & NaN       & 0.964  & 0.943    & 0.931     & 0.930              & 0.970     & 0.981  & 0.970  \\
        seq013\_Pomaria\_1\_int     & lpips  & 0.154  & 0.161       & NaN       & 0.166  & 0.123    & 0.162     & 0.164              & 0.056     & 0.045  & 0.072  \\
        seq014\_Wainscott\_0\_int   & psnr   & 26.011 & 25.966      & NaN       & 29.580 & 32.517   & 30.511    & 30.504             & 31.197    & 31.885 & 31.709 \\
        seq014\_Wainscott\_0\_int   & ssim   & 0.927  & 0.924       & NaN       & 0.925  & 0.955    & 0.951     & 0.951              & 0.952     & 0.969  & 0.958  \\
        seq014\_Wainscott\_0\_int   & lpips  & 0.105  & 0.116       & NaN       & 0.244  & 0.077    & 0.082     & 0.083              & 0.083     & 0.067  & 0.084  \\
        seq015\_Wainscott\_0\_int   & psnr   & 27.257 & 27.191      & NaN       & 32.307 & 30.721   & 28.288    & 28.134             & 34.266    & 32.949 & 35.014 \\
        seq015\_Wainscott\_0\_int   & ssim   & 0.953  & 0.951       & NaN       & 0.962  & 0.955    & 0.942     & 0.942              & 0.976     & 0.975  & 0.980  \\
        seq015\_Wainscott\_0\_int   & lpips  & 0.080  & 0.092       & NaN       & 0.202  & 0.083    & 0.110     & 0.108              & 0.050     & 0.078  & 0.047  \\
        seq016\_Wainscott\_0\_int   & psnr   & 21.953 & 21.660      & 28.364    & 30.205 & 30.414   & 28.915    & 28.710             & 29.746    & 31.965 & 31.096 \\
        seq016\_Wainscott\_0\_int   & ssim   & 0.897  & 0.895       & 0.909     & 0.935  & 0.951    & 0.952     & 0.951              & 0.955     & 0.976  & 0.967  \\
        seq016\_Wainscott\_0\_int   & lpips  & 0.175  & 0.194       & 0.327     & 0.260  & 0.089    & 0.086     & 0.087              & 0.083     & 0.066  & 0.075  \\
        seq017\_Benevolence\_1\_int & psnr   & 26.364 & 26.367      & 27.533    & 30.349 & 29.833   & 29.254    & 26.565             & 31.645    & 28.701 & 28.347 \\
        seq017\_Benevolence\_1\_int & ssim   & 0.927  & 0.920       & 0.897     & 0.923  & 0.937    & 0.933     & 0.887              & 0.948     & 0.970  & 0.958  \\
        seq017\_Benevolence\_1\_int & lpips  & 0.128  & 0.143       & 0.318     & 0.238  & 0.118    & 0.119     & 0.218              & 0.093     & 0.073  & 0.089  \\
        seq018\_Benevolence\_1\_int & psnr   & 28.236 & 24.296      & 32.551    & 34.297 & 34.690   & 33.049    & 33.002             & 34.187    & 34.963 & 33.659 \\
        seq018\_Benevolence\_1\_int & ssim   & 0.918  & 0.809       & 0.911     & 0.936  & 0.951    & 0.953     & 0.952              & 0.958     & 0.976  & 0.966  \\
        seq018\_Benevolence\_1\_int & lpips  & 0.145  & 0.342       & 0.293     & 0.248  & 0.093    & 0.090     & 0.091              & 0.081     & 0.114  & 0.085  \\
        seq019\_Rs\_int             & psnr   & 20.059 & 20.854      & 33.119    & 34.598 & 34.462   & 33.679    & 33.653             & 35.223    & 25.947 & 34.097 \\
        seq019\_Rs\_int             & ssim   & 0.794  & 0.808       & 0.950     & 0.963  & 0.956    & 0.963     & 0.962              & 0.969     & 0.879  & 0.970  \\
        seq019\_Rs\_int             & lpips  & 0.425  & 0.424       & 0.270     & 0.225  & 0.106    & 0.087     & 0.089              & 0.068     & 0.327  & 0.089  \\
        seq020\_Merom\_1\_int       & psnr   & 23.273 & 24.074      & 31.280    & 32.580 & 30.462   & 30.655    & 30.626             & 32.869    & 31.280 & 32.068 \\
        seq020\_Merom\_1\_int       & ssim   & 0.823  & 0.852       & 0.970     & 0.914  & 0.929    & 0.919     & 0.918              & 0.954     & 0.970  & 0.954  \\
        seq020\_Merom\_1\_int       & lpips  & 0.306  & 0.259       & 0.086     & 0.276  & 0.140    & 0.139     & 0.142              & 0.078     & 0.086  & 0.095  \\
        \hline
    \end{tabular}
    } \label{tab:sup_sim_metric}
\end{table*}

\begin{table*}
    [ht]
    \centering
    \setlength{\tabcolsep}{10pt}
    \renewcommand{\arraystretch}{1.0}
    \caption{\textbf{Detailed Quantitative Results on \realdata Dataset}. \ours
    consistently outperforms all other methods in most sequences. Across most sequences,
    \ours maintains high PSNR and SSIM, with low LPIPS, indicating that it
    excels in both numerical image quality and perceptual similarity.}
    \label{tab:supp_interReal} \resizebox{2.0\columnwidth}{!}{%
    \begin{tabular}{l|lccccccc|c}
        \hline
        Dataset             & Metric & NeRF   & Instant-NGP & HyperNeRF & CoNeRF & K-Planes & LiveScene & CoGS   & \ours  \\
        \hline
        seq001\_transformer & psnr   & 20.094 & 20.619      & 24.651    & 27.260 & 26.881   & 30.396    & 31.067 & 31.067 \\
        seq001\_transformer & ssim   & 0.725  & 0.805       & 0.638     & 0.739  & 0.791    & 0.912     & 0.943  & 0.943  \\
        seq001\_transformer & lpips  & 0.182  & 0.167       & 0.495     & 0.355  & 0.185    & 0.060     & 0.060  & 0.060  \\
        seq002\_transformer & psnr   & 20.093 & 20.028      & 24.433    & 26.917 & 26.232   & 29.706    & 30.513 & 30.513 \\
        seq002\_transformer & ssim   & 0.736  & 0.778       & 0.635     & 0.732  & 0.763    & 0.899     & 0.938  & 0.938  \\
        seq002\_transformer & lpips  & 0.210  & 0.196       & 0.477     & 0.357  & 0.223    & 0.069     & 0.062  & 0.062  \\
        seq003\_door        & psnr   & 20.001 & 20.652      & 27.144    & 29.850 & 29.278   & 32.709    & 31.998 & 31.998 \\
        seq003\_door        & ssim   & 0.785  & 0.831       & 0.878     & 0.922  & 0.920    & 0.960     & 0.962  & 0.962  \\
        seq003\_door        & lpips  & 0.250  & 0.250       & 0.316     & 0.231  & 0.101    & 0.044     & 0.071  & 0.071  \\
        seq004\_dog         & psnr   & 20.044 & 20.206      & 25.691    & 28.567 & 30.350   & 32.519    & 32.455 & 33.555 \\
        seq004\_dog         & ssim   & 0.723  & 0.819       & 0.730     & 0.815  & 0.894    & 0.943     & 0.950  & 0.960  \\
        seq004\_dog         & lpips  & 0.196  & 0.178       & 0.435     & 0.324  & 0.107    & 0.049     & 0.074  & 0.063  \\
        seq005\_sit         & psnr   & 21.558 & 24.211      & 24.944    & 26.252 & 27.970   & 30.161    & 27.169 & 30.236 \\
        seq005\_sit         & ssim   & 0.480  & 0.727       & 0.573     & 0.633  & 0.773    & 0.886     & 0.767  & 0.912  \\
        seq005\_sit         & lpips  & 0.178  & 0.236       & 0.543     & 0.463  & 0.207    & 0.084     & 0.232  & 0.098  \\
        seq006\_stand       & psnr   & 23.109 & 24.483      & 24.833    & 26.159 & 27.285   & 29.400    & 31.442 & 30.489 \\
        seq006\_stand       & ssim   & 0.643  & 0.699       & 0.574     & 0.627  & 0.736    & 0.868     & 0.919  & 0.913  \\
        seq006\_stand       & lpips  & 0.123  & 0.260       & 0.538     & 0.470  & 0.237    & 0.089     & 0.104  & 0.092  \\
        seq007\_flower      & psnr   & 21.150 & 21.813      & 25.334    & 26.854 & 26.545   & 28.208    & 28.435 & 28.435 \\
        seq007\_flower      & ssim   & 0.721  & 0.747       & 0.712     & 0.748  & 0.759    & 0.844     & 0.893  & 0.893  \\
        seq007\_flower      & lpips  & 0.302  & 0.319       & 0.489     & 0.425  & 0.321    & 0.188     & 0.165  & 0.165  \\
        seq008\_office      & psnr   & 21.187 & 21.474      & 25.188    & 26.040 & 26.309   & 28.663    & 27.510 & 27.620 \\
        seq008\_office      & ssim   & 0.735  & 0.743       & 0.714     & 0.720  & 0.754    & 0.848     & 0.897  & 0.872  \\
        seq008\_office      & lpips  & 0.371  & 0.358       & 0.545     & 0.520  & 0.341    & 0.181     & 0.138  & 0.181  \\
        \hline
    \end{tabular}
    } \label{tab:sup_real_metric}
\end{table*}

\begin{table*}
    [ht]
    \vspace{-1ex}
    \centering
    \setlength{\tabcolsep}{3pt}
    \renewcommand{\arraystretch}{0.6}
    \caption{\textbf{Quantitative results on DyNeRF datasets.} \ours ranks first
    in PSNR on 4/6 DyNeRF sences.}
    \vspace{-1ex}
    \resizebox{2\columnwidth}{!}{
    \begin{tabular}{lccccccc}
        \toprule Method                          & Coffee Martini & Cook Spinach   & Cut Beef       & Flame Salmon   & Flame Steak    & Sear Steak     & Mean           \\
        \midrule HexPlane~\cite{Cao2023HexPlane} & ---            & 32.04          & 32.55          & 29.47          & 32.08          & 32.39          & 31.70          \\
        K-Planes~\cite{kplanes_2023}             & 29.99          & 32.60          & 31.82          & 30.44          & 32.38          & 32.52          & 31.63          \\
        MixVoxels\cite{wang2023mixed}            & 29.36          & 31.61          & 31.30          & 29.92          & 31.21          & 31.43          & 30.80          \\
        NeRFPlayer~\cite{song2023nerfplayer}     & 31.53          & 30.56          & 29.35          & 31.65          & 31.93          & 29.12          & 30.69          \\
        HyperReel~\cite{attal2023hyperreel}      & 28.37          & 32.30          & 32.92          & 28.26          & 32.20          & 32.57          & 31.10          \\
        4DGS~\cite{wu20234dgaussians}            & 27.34          & 32.46          & 32.90          & 29.20          & 32.51          & 32.49          & 31.15          \\
        RT-4DGS~\cite{yang2023real}              & 28.33          & 32.93          & 33.85          & 29.38          & 34.03          & 33.51          & 32.01          \\
        GaussianFlow~\cite{gao2024gaussianflow}  & 28.42          & \textbf{33.68} & 34.12          & 29.36          & 34.22          & \textbf{34.00} & 32.30          \\
        \ours (Ours)                             & \textbf{28.53} & 33.36          & \textbf{34.33} & \textbf{29.58} & \textbf{34.29} & 33.83          & \textbf{32.32} \\
        \bottomrule
    \end{tabular}
    } \label{tab:supp_dynerf}
\end{table*}

\boldparagraph{View Synthesis Quality Comparison on \simdata and \realdata dataset}
We present detailed quantitative results on the \simdata and \realdata datasets
in~\cref{tab:sup_sim_metric} and~\cref{tab:sup_real_metric}, respectively. Our
method demonstrates significant advantages on both the \#easy and \#medium
subsets of the \simdata dataset. Additionally, it achieves notable scores on the
\#medium subset of the \realdata dataset. A multitude of metrics indicate that
our model excels in rendering on both simulated and real datasets, underscoring its
superiority. While the metric improvements may be modest compared to current
SOTA NeRF methods, our approach offers a substantial advantage by introducing a novel
guidance-free training paradigm that significantly reduces the label requirements,
thereby enhancing its real-world applicability. We report scores as NaN if the model
fails to converge or runs out of memory during training multiple times.

\begin{figure*}[t]
    \vspace{-1ex}
    \begin{center}
        \includegraphics[width=1\linewidth]{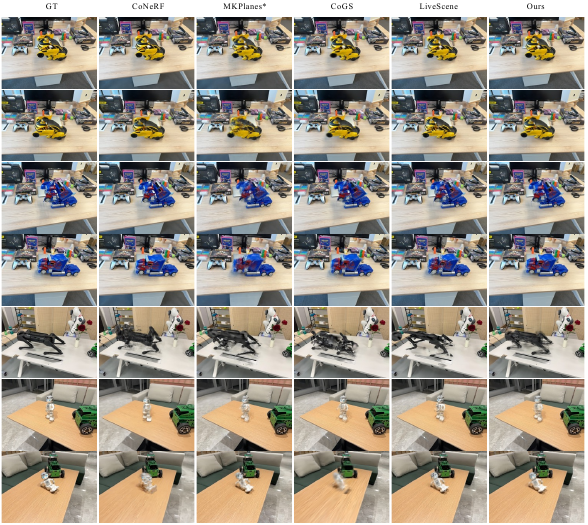}
    \end{center}
    \vspace{-3ex}
    \caption{\textbf{View Synthesis Visualization on \realdata Dataset}. We
    compare our method with SOTA methods on RGB rendering across real scenes. \ours
    obtained more detailed and accurate representations of the objects. While
    other methods fail to capture the object's shape and cause significant
    artifacts.}
    \label{fig:supp_render_real}
    \vspace{-2ex}
\end{figure*}

\begin{figure*}[h]
    \begin{center}
        \includegraphics[width=0.98\linewidth]{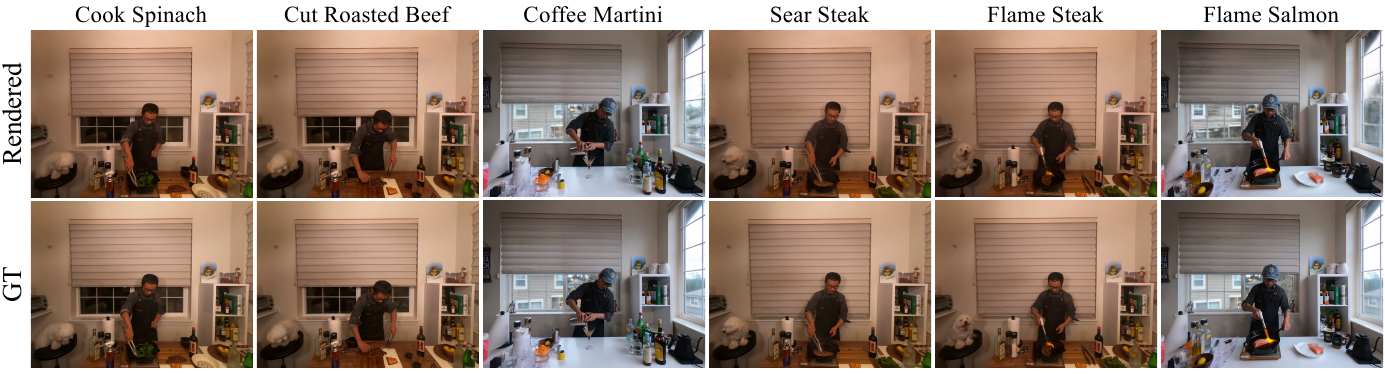}
    \end{center}
    \vspace{-2ex}
    \caption{View Synthesis Visualization on DyNeRF Dataset.}
    \label{fig:supp_dynerf_vis}
\end{figure*}

\begin{figure*}[t]
    \vspace{-1ex}
    \begin{center}
        \includegraphics[width=1\linewidth]{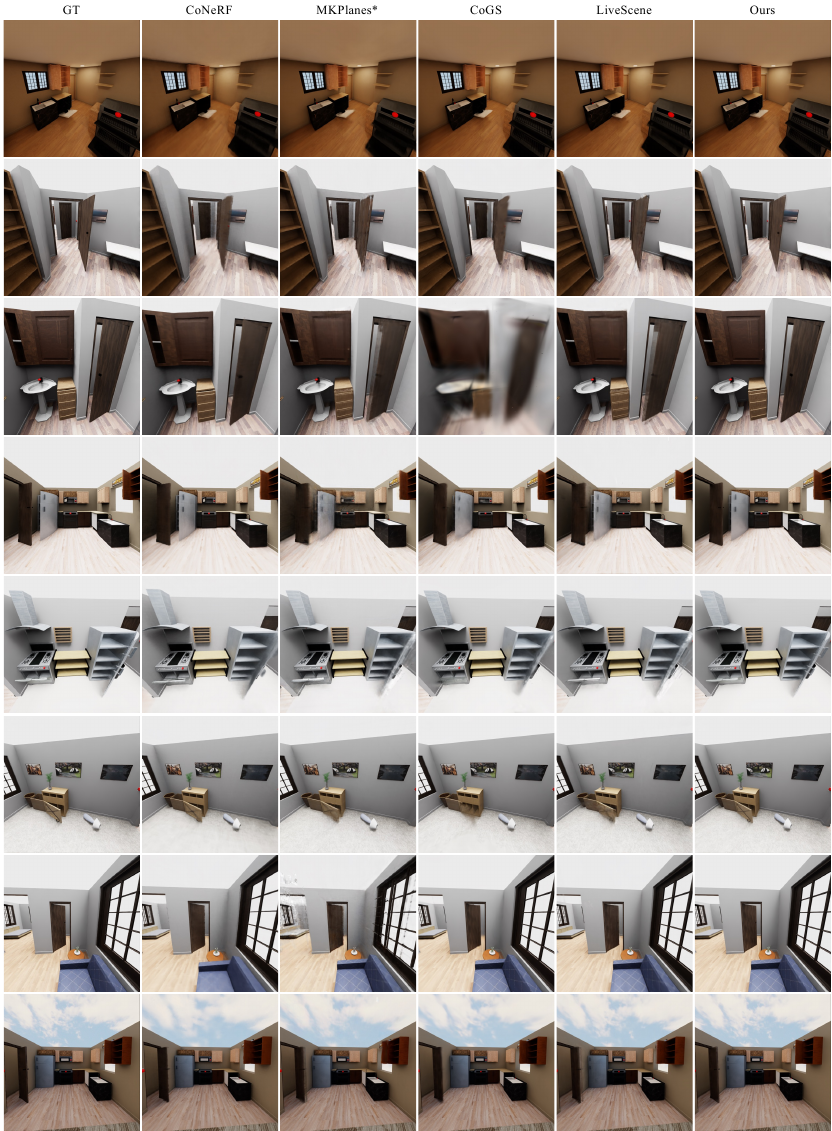}
    \end{center}
    \vspace{-1ex}
    \caption{\textbf{View Synthesis Visualization on \simdata Dataset}. Compared
    with the other methods, \ours reconstructs clear and accurate object shapes and
    textures.}
    \label{fig:supp_render_sim}
    \vspace{-2ex}
\end{figure*}

\boldparagraph{More Detailed Rendering Comparison} We show additional visual
comparisons in~\cref{fig:supp_render_real}, \cref{fig:supp_render_sim},
showcasing our method's superior performance on the \simdata and \realdata
datasets. Our approach excels in reconstructing detailed and accurate object representations.
Notably, our method generates more accurate object shapes and background textures
compared to existing approaches. We also provide a visualization of DyNeRF
dataset in~\cref{fig:supp_dynerf_vis} to show the rendering quality in 4D
dynamic scene.

\begin{figure*}[h]
    \vspace{-1ex}
    \begin{center}
        \includegraphics[width=1\linewidth]{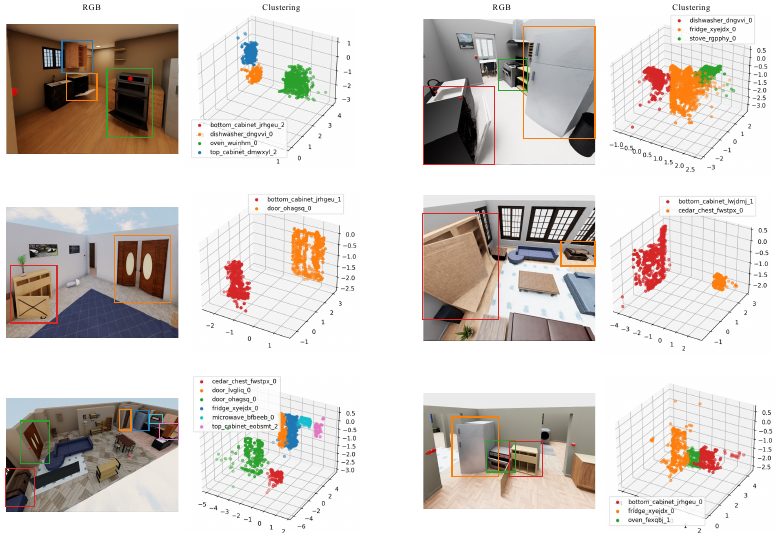}
    \end{center}
    \vspace{-1ex}
    \caption{\textbf{Visualization of HDBSCAN clustering}. After successfully
    training the 4D Gaussian field, we apply HDBSCAN and the interaction flow to
    identify the key Gaussian spheres corresponding to the controllable objects.}
    \label{fig:supp_tracking}
    \vspace{-2ex}
\end{figure*}

\boldparagraph{More Detailed Clustering Visualization} ~\cref{fig:supp_tracking}
illustrates the clustering results of our method across various scenarios. As
demonstrated, the majority of Gaussian clusters are accurately grouped around controllable
entities, particularly in relation to the moving components. This can be attributed
to the successful decoupling of the interaction flow, a feature that enables the
Gaussian clusters to concentrate more effectively on the motion rendering.

\begin{figure*}[t]
    \vspace{-1ex}
    \begin{center}
        \includegraphics[width=0.5\linewidth]{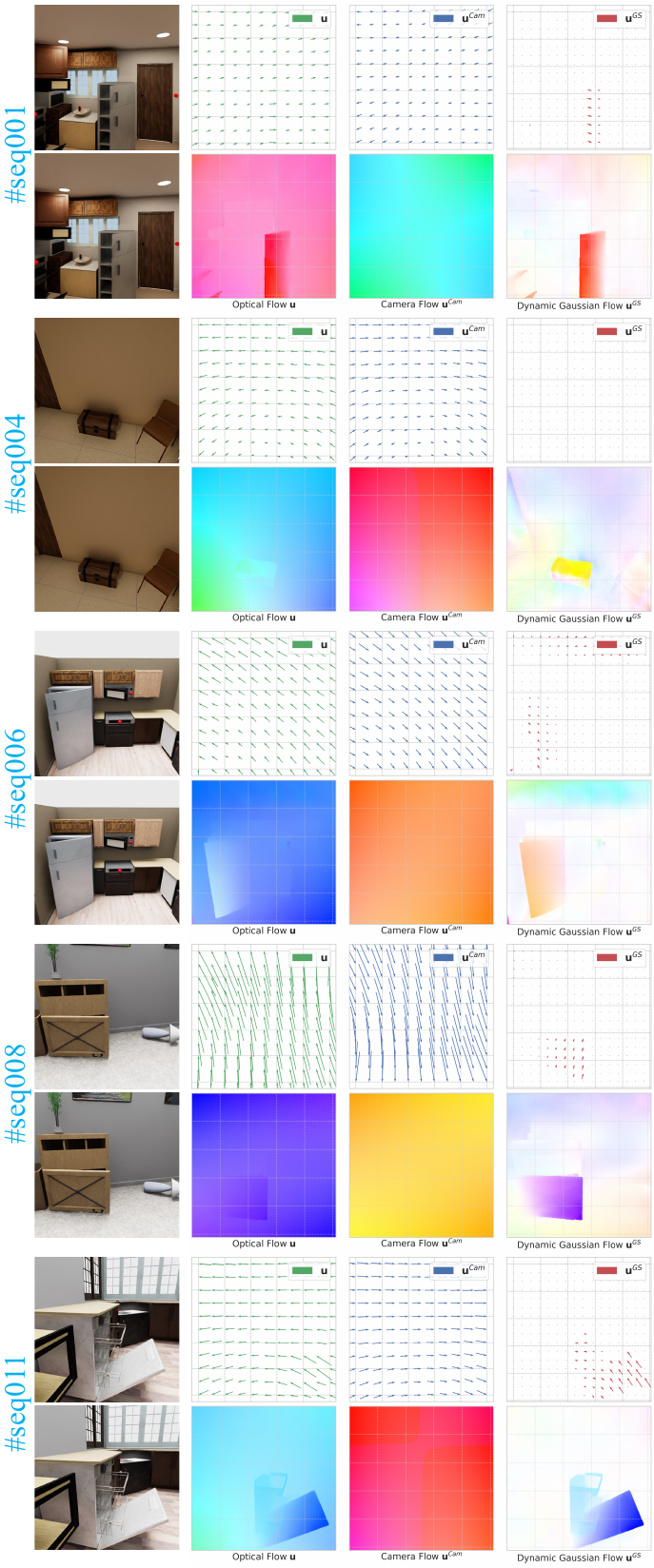}
    \end{center}
    \vspace{-1ex}
    \caption{More illustrations of dynamic Gaussian flow map under dynamic
    scenes of \simdata. For dynamic scenes with interactive objects and complex camera
    motions (translation and rotation), the dynamic Gaussian flow map will
    highlight interactive 3D Gaussians, and demonstrate the effectiveness of proposed
    Dynamic Gaussian Flow derivatives in~\cref{eq:supp_gaussian_flow_analysis}.}
    \label{fig:supp_flowmap_sim}
    \vspace{-2ex}
\end{figure*}

\begin{figure*}[h]
    \vspace{-1ex}
    \begin{center}
        \includegraphics[width=0.98\linewidth]{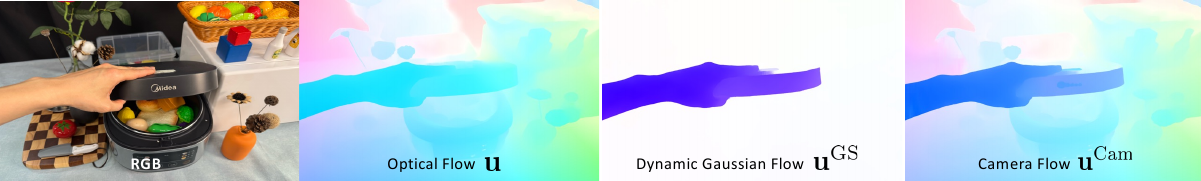}
    \end{center}
    \vspace{-1ex}
    \caption{More illustrations of dynamic Gaussian flow map under dynamic
    scenes of self-captured data. the rice-cooker lid is lifted while the camera
    translates,the estimated camera flow captures the egomotion and the dynamic Gaussian
    flow isolates the hand motion. The clean separation empirically validates
    the optimisation objective of proposed Dynamic Gaussian Flow derivatives in~\cref{eq:supp_gaussian_flow_analysis}.}
    \label{fig:supp_flowmap_real}
    \vspace{-2ex}
\end{figure*}

\boldparagraph{More illustrations of dynamic Gaussian flow map} We provide a more
detailed visualization of highlighting dynamic Gaussian capabilities in~\cref{fig:supp_flowmap_real,fig:supp_flowmap_sim}.
The experimental results show that, despite the presence of complex camera motion
and interactive body motion, the proposed approach successfully decouples the
Gaussian dynamics, producing accurate and detailed flow maps. Notably, objects exhibiting
complex topological structure changes, such as boxes or dishwashers, can be effectively
isolated. This outcome substantiates the efficacy and unsupervised exploration
capabilities of the proposed method for interactive Gaussian discovery.

\begin{figure*}[h]
    \vspace{-3ex}
    \begin{center}
        \includegraphics[width=0.98\linewidth]{
            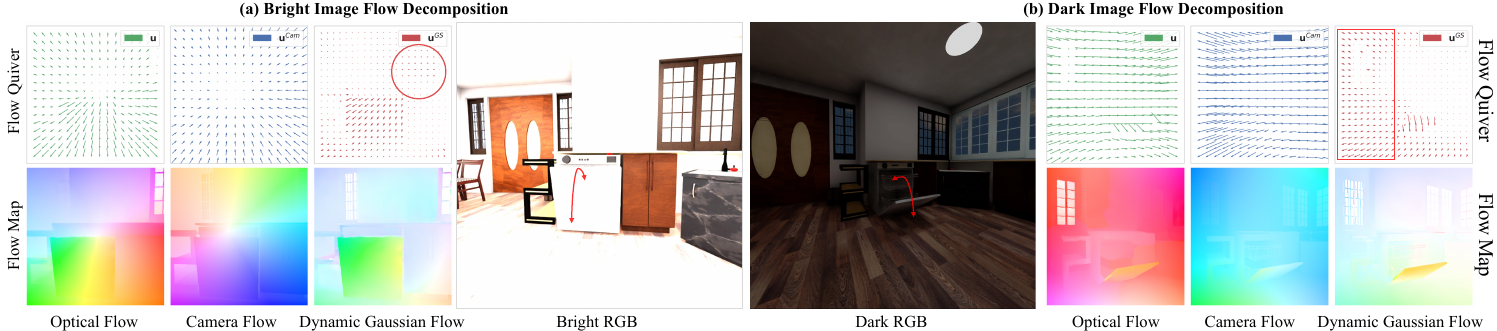
        }
    \end{center}
    \vspace{-2ex}
    \caption{Failure cases due to excessively intense or insufficient lighting.}
    \label{fig:supp_flow_failure}
\end{figure*}

\boldparagraph{More illustrations of failure cases} ~\cref{fig:supp_flow_failure}
shows that due to uneven lighting, the flow estimator overestimates the environmental
flow, resulting in corresponding high-brightness regions in non-moving areas. The
inaccuracy of flow estimation affects the clustering results and ultimately influences
the final control process.

\begin{figure*}[h]
    \vspace{-3ex}
    \begin{center}
        \includegraphics[width=0.98\linewidth]{
            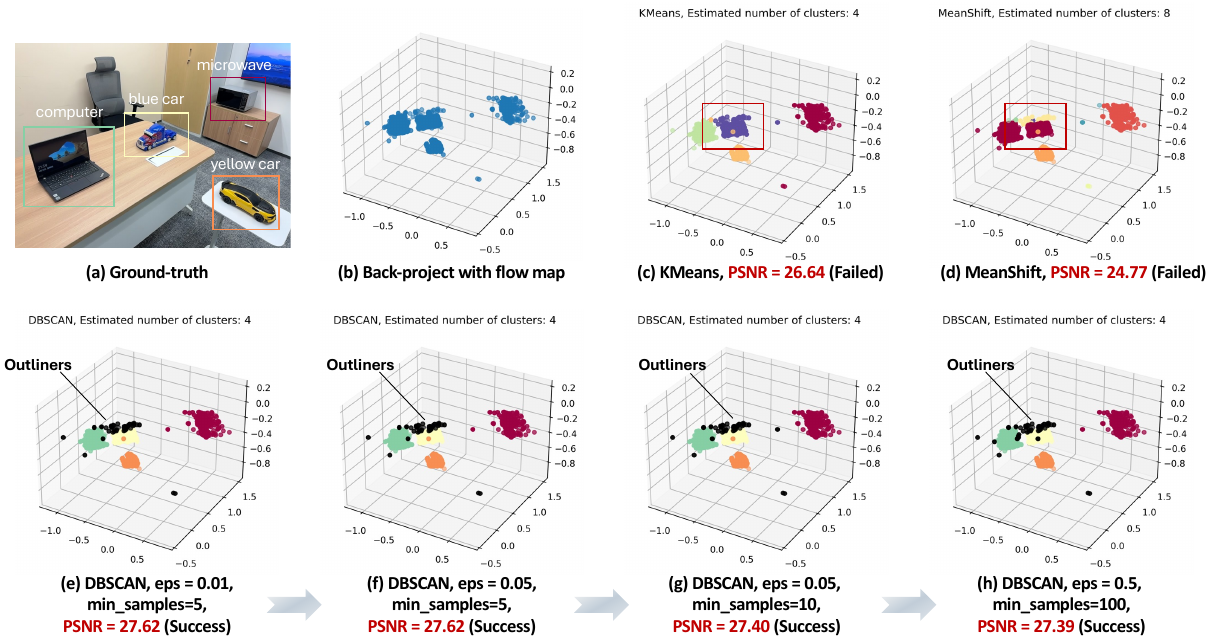
        }
    \end{center}
    \vspace{-2ex}
    \caption{Comparison of clustering results among KMeans, MeanShift and HDBSCAN
    with varying parameters on \#seq008 of \realdata.}
    \label{fig:supp_cluster_sim}
\end{figure*}

\clearpage
\end{document}